\documentclass[sigconf]{acmart}
\AtBeginDocument{%
  }

\usepackage{placeins}
\usepackage{balance}

\usepackage{booktabs}
\usepackage{amsmath} 
\usepackage{todonotes}

\usepackage{bbm}
\usepackage{enumitem}

\usepackage{listings}
\usepackage{xcolor}

\usepackage{graphicx}
\usepackage{tabularx}
\usepackage{xcolor} % For coloring text
\definecolor{lightblue}{rgb}{.50,.90,0.51}
\definecolor{tri}{rgb}{.25,.88,.82}
\definecolor{lilac}{rgb}{0.85,0.64,0.85}
\definecolor{atomictangerine}{rgb}{1.0, 0.6, 0.4}
\copyrightyear{2026}
\acmYear{2026}
\setcopyright{cc}
\setcctype{by}
\acmConference[WWW Companion '26]{Companion Proceedings of the ACM Web Conference 2026}{April 13--17, 2026}{Dubai, United Arab Emirates}
\acmBooktitle{Companion Proceedings of the ACM Web Conference 2026 (WWW Companion '26), April 13--17, 2026, Dubai, United Arab Emirates}
\acmPrice{}
\acmDOI{10.1145/3774905.3795465}
\acmISBN{979-8-4007-2308-7/2026/04}

\begin{document}

%%
%% The "title" command has an optional parameter,
%% allowing the author to define a "short title" to be used in page headers.
% \title{MemeReason: Toward Robust Reasoning for Propaganda and Hate in Multimodal Memes}
\title{Can Thinking Models Think to Detect Hateful Memes?
}

% \author{
% Mohamed Bayan Kmainasi\textsuperscript{\rm 1},
% Mucahid Kutlu\textsuperscript{\rm 1}, 
% Ali Ezzat Shahroor\textsuperscript{\rm 2},\\ 
% Abul Hasnat\textsuperscript{\rm 3},\textsuperscript{\rm 4} 
% Firoj Alam\textsuperscript{\rm 2}
% }
% \affiliation{%
%   \textsuperscript{\rm 1}Qatar University, Qatar, 
%   \textsuperscript{\rm 2}Qatar Computing Research Institute, Qatar, \\  
%   \textsuperscript{\rm 3}Blackbird.AI, USA,
%   \textsuperscript{\rm 4}APAVI.AI, France\\
% \{mk2314890, mucahidkutlu\}@qu.edu.qa, \{fialam, alsh34060\}@hbku.edu.qa, mhasnat@gmail.com
% }

% \author{Mohamed Bayan Kmainasi}
% \affiliation{%
%   \institution{Qatar University}
%   \city{Doha}
%   \country{Qatar}
% }
% \email{mk2314890@qu.edu.qa}

% \author{Mucahid Kutlu}
% \affiliation{%
%   \institution{Qatar University}
%   \city{Doha}
%   \country{Qatar}
% }
% \email{mucahidkutlu@qu.edu.qa}

% \author{Ali Ezzat Shahroor}
% \affiliation{%
%   \institution{Qatar Computing Research Institute, HBKU}
%   \city{Doha}
%   \country{Qatar}
% }
% \email{alsh34060@hbku.edu.qa}

% \author{Firoj Alam}
% \affiliation{%
%   \institution{Qatar Computing Research Institute, HBKU}
%   \city{Doha}
%   \country{Qatar}
% }
% \email{fialam@hbku.edu.qa}

% \author{Abul Hasnat}
% \affiliation{%
%   \institution{Blackbird.AI}
%   \city{New York}
%   \state{NY}
%   \country{USA}
% }
% \affiliation{%
%   \institution{APAVI.AI}
%   \city{Paris}
%   \country{France}
% }
% \email{mhasnat@gmail.com}

\author{Mohamed Bayan Kmainasi}
\orcid{0009-0005-4115-6057}
\affiliation{\institution{Qatar University}\city{Doha}\country{Qatar}}
\email{mk2314890@qu.edu.qa}

\author{Mucahid Kutlu}
\orcid{0000-0002-5660-4992}
\affiliation{\institution{Qatar University}\city{Doha}\country{Qatar}}
\email{mucahidkutlu@qu.edu.qa}

\author{Ali Ezzat Shahroor}
\orcid{0009-0004-3918-7471}
\affiliation{\institution{Qatar Computing Research Institute}\city{Doha}\country{Qatar}}
\email{alsh34060@hbku.edu.qa}

\author{Abul Hasnat}
\orcid{0000-0003-2748-8221}
\affiliation{\institution{APAVI.AI}\country{France}}
\affiliation{\institution{Blackbird.AI}\country{USA}}
\email{mhasnat@gmail.com}

\author{Firoj Alam}
\orcid{0000-0001-7172-1997}
\affiliation{\institution{Qatar Computing Research Institute}\city{Doha}\country{Qatar}}
\email{fialam@hbku.edu.qa}

%%
%% By default, the full list of authors will be used in the page
%% headers. Often, this list is too long, and will overlap
%% other information printed in the page headers. This command allows
%% the author to define a more concise list
%% of authors' names for this purpose.
\renewcommand{\shortauthors}{Mohamed Bayan Kmainasi, Mucahid Kutlu, Ali Ezzat Shahroor, Abul Hasnat, \& Firoj Alam}

%%
%% The abstract is a short summary of the work to be presented in the
%% article.
\begin{abstract}
Hateful memes often require compositional multimodal reasoning: the image and text may appear benign in isolation, yet their interaction conveys harmful intent. Although thinking-based multimodal large language models (MLLMs) have recently advanced vision–language understanding, their capabilities remain underexplored for hateful meme analysis. We propose a reinforcement learning–based post-training framework that improves reasoning in thinking-based MLLMs via task-specific rewards and a novel Group Relative Policy Optimization (GRPO) objective. Concretely, we \textit{(i)} conduct a systematic empirical study of off-the-shelf MLLMs for hateful meme understanding, \textit{(ii)} extend an existing hateful meme dataset by generating weakly/pseudo-supervised chain-of-thought (CoT) rationales via distillation, and \textit{(iii)} introduce a GRPO-based objective that jointly optimizes meme classification and explanation quality to encourage fine-grained step-by-step reasoning. Experiments on the \textit{Hateful Memes} benchmark show that our approach achieves state-of-the-art results, improving accuracy and F1 by approximately 1\% and explanation quality by approximately 3\%. We will publicly release our code, data extensions, and evaluation resources to support reproducibility.\footnote{\url{https://github.com/MohamedBayan/MemeReason}}
\end{abstract}



% \vspace{0.1cm}

%%
%% The code below is generated by the tool at http://dl.acm.org/ccs.cfm.
%% Please copy and paste the code instead of the example below.
%%
% \begin{CCSXML}
% <ccs2012>
%  <concept>
%   <concept_id>00000000.0000000.0000000</concept_id>
%   <concept_desc>Do Not Use This Code, Generate the Correct Terms for Your Paper</concept_desc>
%   <concept_significance>500</concept_significance>
%  </concept>
%  <concept>
%   <concept_id>00000000.00000000.00000000</concept_id>
%   <concept_desc>Do Not Use This Code, Generate the Correct Terms for Your Paper</concept_desc>
%   <concept_significance>300</concept_significance>
%  </concept>
%  <concept>
%   <concept_id>00000000.00000000.00000000</concept_id>
%   <concept_desc>Do Not Use This Code, Generate the Correct Terms for Your Paper</concept_desc>
%   <concept_significance>100</concept_significance>
%  </concept>
%  <concept>
%   <concept_id>00000000.00000000.00000000</concept_id>
%   <concept_desc>Do Not Use This Code, Generate the Correct Terms for Your Paper</concept_desc>
%   <concept_significance>100</concept_significance>
%  </concept>
% </ccs2012>
% \end{CCSXML}

% \ccsdesc[500]{Do Not Use This Code~Generate the Correct Terms for Your Paper}
% \ccsdesc[300]{Do Not Use This Code~Generate the Correct Terms for Your Paper}
% \ccsdesc{Do Not Use This Code~Generate the Correct Terms for Your Paper}
% \ccsdesc[100]{Do Not Use This Code~Generate the Correct Terms for Your Paper}

%%
%% Keywords. The author(s) should pick words that accurately describe
%% the work being presented. Separate the keywords with commas.
%% A "teaser" image appears between the author and affiliation
%% information and the body of the document, and typically spans the
%% page.

% \begin{teaserfigure}
%   \includegraphics[width=\textwidth]{sampleteaser}
%   \caption{Seattle Mariners at Spring Training, 2010.}
%   \Description{Enjoying the baseball game from the third-base
%   seats. Ichiro Suzuki preparing to bat.}
%   \label{fig:teaser}
% \end{teaserfigure}

% \received{20 February 2007}
% \received[revised]{12 March 2009}
% \received[accepted]{5 June 2009}

%%
%% This command processes the author and affiliation and title
%% information and builds the first part of the formatted document.

\maketitle

% \footnotesize{\textcolor{red}{WARNING: This paper contains examples which may be disturbing to the reader}}

% {\color{red}\textbf{Content Notice}: \textit{This paper contains examples of hateful and discriminatory content that may be upsetting to some readers.}}

\section{Introduction}
%% Problem scope
Memes, which blend images and text with humor and cultural references, are classic examples of \textit{multimodal} content. They have become a predominant form of communication on social media, rapidly shaping collective discourse \cite{pandiani2025toxic, alafnan2025role}. While often harmless, they exhibit a dual nature and can be exploited to spread hate speech, disinformation, and propaganda \cite{sharma2022detecting}. The use of humor and irony in hateful memes may trivialize toxic content, potentially normalizing hostile views \cite{schmid2025humorous, pandiani2025toxic}. 
Furthermore, memes serve as powerful tools for political communication, effectively conveying complex ideas through simple, emotionally resonant visuals that enable persuasion and manipulation \cite{alafnan2025role, mihuailescu2024never}. 
Given the pervasive role of social media, timely and accurate detection and moderation of hateful content are therefore crucial to ensuring safer online environments. Motivated by these challenges, \textit{this research aims to improve hateful meme detection by leveraging recent advances in multimodal large language models (MLLMs)}.

%% Recent advancement on MLLMs
Recently, MLLMs have significantly advanced vision–language understanding, demonstrating strong generalization capabilities across a wide range of multimodal tasks \cite{sun2025multi, wu2024visionllm}. 
Beyond strong representation learning, these models increasingly exhibit emergent abilities such as instruction following, cross-modal grounding, and robust transfer to tasks with limited task-specific supervision, making them attractive foundations for challenging content understanding problems.
\textit{Thinking-based MLLMs} represent the latest advances in multimodal reasoning, exhibiting an enhanced ability to integrate visual and textual information through deliberate multi-step inference \cite{yang2025qwen3}.
This paradigm is particularly relevant for memes, where meaning is often implicit and depends on image-text interaction rather than unimodal cues.
The training of such models emphasizes chain-of-thought (CoT) supervision, in which intermediate reasoning steps are explicitly modeled, and is further optimized through reinforcement learning (RL) techniques \cite{ deepseekai2025deepseekr1incentivizingreasoningcapability}.
%wei2025sftsecondrlupt, sun-etal-2024-aligning, yu2024rlhf, ahn2024tuning, yu2025aligning , cao2025survey, tie2025survey}.
%
%

\begin{figure*}[t]
    \centering
    \includegraphics[width=0.83\textwidth]{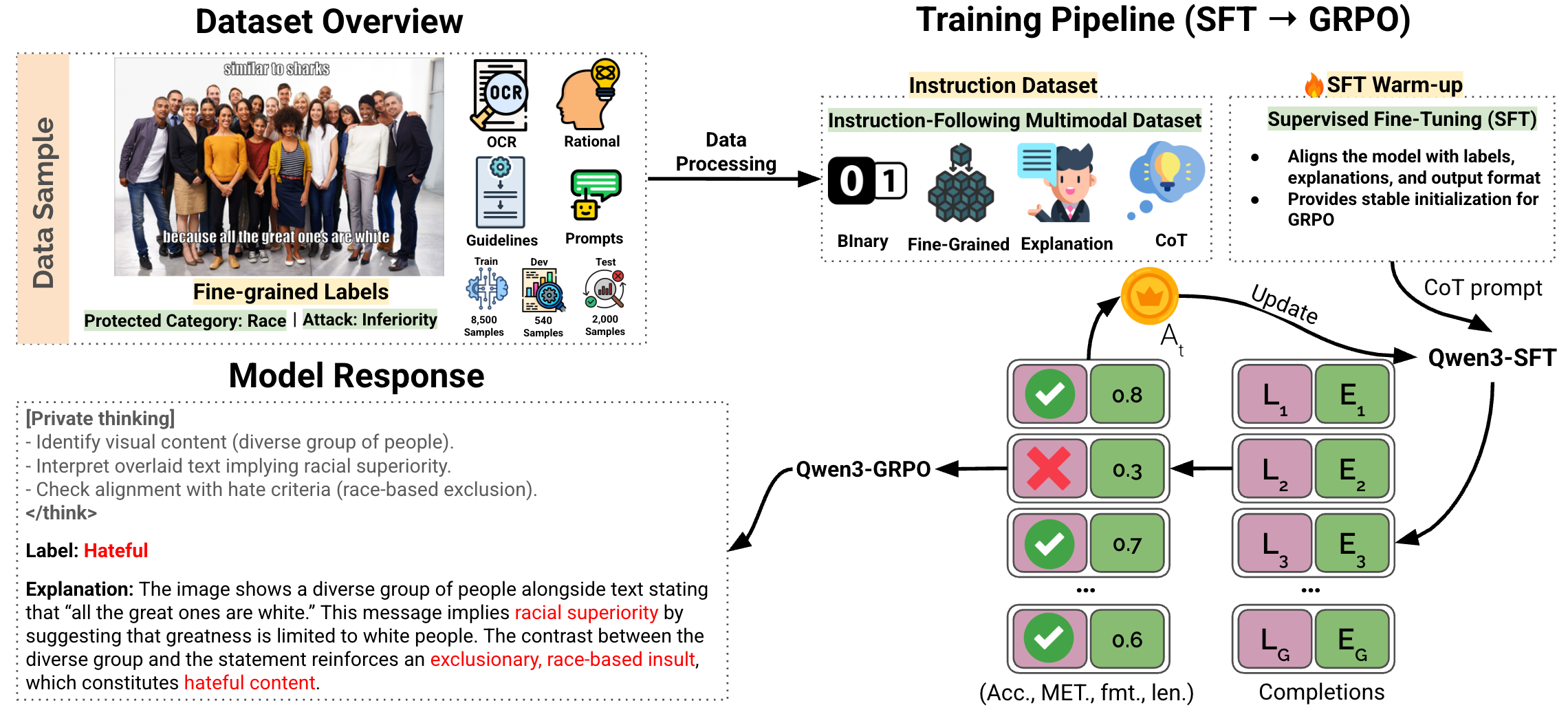}
    \caption{
    Overview of the proposed hateful meme analysis framework.
    Starting from raw memes, we derive binary and fine-grained supervision (protected category and attack type), OCR text, and guidelines to construct instruction-following datasets.
    Weak supervision from a strong MLLM is used to distill step-by-step CoT rationales.
    The model is trained via a two-stage post-training pipeline consisting of SFT warm-up followed by GRPO-based reinforcement learning, jointly optimizing classification accuracy and explanation quality.
    During inference, multiple candidate label–explanation pairs are generated and scored to select the final prediction.
    Abbreviations: \textbf{Acc.} = accuracy; \textbf{MET} = METEOR; \textbf{Len} = explanation length; \textbf{Fmt} = format compliance.
    }
    \label{fig:overview}
\end{figure*}

%% What has been done and what we want to propose
Significant progress has been achieved through diverse approaches for meme understanding, including agentic reasoning frameworks \cite{lu2025having, liu2025mind}, few-shot adaptation \cite{cao2024modularized, heebridging}, and Supervised fine-tuning (SFT) on multimodal datasets \cite{alam-etal-2024-armeme, kmainasi2025memeintel}. However, only very recent or concurrent work \cite{mei2025expo} has begun to explore chain-of-thought, supervised training, and reinforcement learning for meme analysis. 
In addition, thinking-based MLLMs have not yet been explored for hateful meme analysis. 

To address this research gap, and as illustrated in Figure~\ref{fig:overview}, \textit{we investigate this direction and develop an RL-trained approach grounded in the CoT-supervised reasoning paradigm. Specifically, we first study and analyze the impact of CoT on meme analysis using pre-trained thinking-based MLLMs, and subsequently develop a novel method (see Section \ref{sec_methodology} for more details) to fine-tune thinking-based MLLMs for hateful meme analysis.}

Our contributions are summarized as follows:
\begin{itemize}[noitemsep,topsep=0pt,labelsep=.5em] 
    \item \textbf{Empirical study of CoT reasoning:}
    We evaluate recent pre-trained MLLMs for hateful meme understanding under zero-shot and CoT prompting, jointly assessing classification and explanation generation to identify a strong base model.
    \item \textbf{Extended with step-by-step reasoning:} 
    % \todo{"Dataset Extension with step-by-step reasoning" seems to me better.}
    We extend an existing hateful meme dataset 
    % \todo{I think we should give the name of the dataset} 
    with distilled step-by-step CoT rationales, obtained via weak/pseudo supervision from a strong commercial MLLM. 
    % \todo{I think we can give the name of the MLLM}
    \item \textbf{GRPO optimization objective:}
    We introduce a GRPO-based objective for meme classification with meteor-based reward objective that jointly optimizes label correctness and explanation quality for hateful meme analysis.
    \item \textbf{Two-stage GRPO fine-tuning:}
    We develop a post-training pipeline with SFT warm-up followed by GRPO-based reinforcement learning for explainable hateful meme detection.
\end{itemize}

Our findings can be summarized as follows: \textit{(a)} thinking-based pre-trained MLLMs consistently benefit from CoT prompting; \textit{(b)} incorporating fine-grained supervision and distilled CoT rationales further improves both classification and explanation quality under our GRPO-based post-training; and \textit{(c)} the proposed GRPO objective and RL-based training method achieve state-of-the-art results for hateful meme classification and explanation.

\section{Related Work}
\subsection{Multimodal hateful meme detection}
Early meme detection systems relied on unimodal cues, such as OCR-based text features combined with traditional machine learning algorithms \cite{amalia2018meme, boinepelli2020sis}, CNN-based image classifiers and image--text fusion methods \cite{shrestha2020nlp_uiowa}. However, these approaches often struggled to capture the nuanced, context-dependent meanings arising from complex image--text interactions.

Most existing research on meme moderation has focused on binary classification, distinguishing between categories such as hateful and non-hateful content \cite{cao2023pro}. Several benchmark datasets have been developed to support this line of research, including the 
% Facebook Hateful Memes (FHM) 
Hateful Meme dataset introduced by \citet{kiela2021hatefulmemeschallengedetecting} and HarMeme, which targets COVID-19 and U.S. political memes \cite{pramanick2021momenta}.
%For Arabic content, the only available resource is ArMeme \cite{alam-etal-2024-armeme}.

Only a limited number of works have explored fine-grained annotation. For example, \citet{mathias-etal-2021-findings} re-labeled Hateful Memes with fine-grained attributes based on protected categories and attack types, while \citet{fersini-etal-2022-semeval} introduced MAMI, a dataset for misogynistic meme detection that supports both binary and subtype-level classification according to the type of misogyny expressed.

More recently, several studies have moved beyond detection toward explainability, taxonomic labeling, and enhanced multimodal reasoning. For instance, \citet{grasso2024kermit} augmented multimodal classification with knowledge-graph reasoning using ConceptNet to enrich contextual understanding. The Mod-Hate framework \cite{cao2024modularized} addressed few-shot hateful meme detection by modularizing LLM components, while \citet{lin2024towards} proposed a multimodal debate framework in which multiple LLM agents exchange rationales before reaching a final decision. In a related effort, \citet{kmainasi2025memeintel} extended FHM by incorporating human-verified rationales, enabling supervised training of explainable MLLMs.

\textcolor{black}{
Despite these advances, systematic investigations to enhance CoT reasoning over multimodal inputs remain limited. Existing work has explored CoT prompting or structured reasoning in specific settings, such as multi-hop reasoning for misogynistic meme classification \cite{kumari2024m3hop}, multimodal metaphor detection \cite{xu2024exploring},
% (Xu et al., 2024), 
and distilled abductive reasoning for harmful meme detection \cite{lin-etal-2023-beneath}.
% (Lin et al., 2023). 
However, the use of RL/GRPO for multimodal hateful meme detection remains largely underexplored. To the best of our knowledge, only very recent or concurrent work~\cite{mei2025expo} has begun to investigate GRPO in the context of meme moderation. Prior research has predominantly relied on supervised or prompting-based paradigms, leaving the potential of RL for improving multimodal reasoning in this domain insufficiently examined.
}

\subsection{RL for MLLMs}
RL is a core paradigm in machine learning where the agent learns to make decisions by interacting with an environment and receiving rewards \cite{kaelbling1996reinforcementlearningsurvey}. Recent advances in MLLMs have leveraged RL for post-training with dynamic feedback, where reward signals guide model behavior \cite{wu2025sailing}. Methods such as Reinforcement Learning from Human Feedback (RLHF) have been extensively employed to align and refine the behavior of LLMs \cite{ouyang2022training, kaufmann2024survey}.

These methods typically utilize optimization algorithms such as Proximal Policy Optimization (PPO) \cite{schulman2017proximal} and Direct Preference Optimization (DPO) \cite{rafailov2023direct} to adjust model parameters toward maximizing the expected reward signal. More recently, the DeepSeek-R1 model introduced GRPO \cite{deepseekai2025deepseekr1incentivizingreasoningcapability}, achieving state-of-the-art reasoning performance. Unlike traditional RLHF approaches that rely on learned evaluation models, GRPO employs a rule-based reward that explicitly encourages the generation of high-quality chains of thoughts and favors trajectories exhibiting superior reasoning quality.
\textcolor{black}{GRPO has been effectively applied across a range of domains, including mathematical reasoning \cite{shao2024deepseekmathpushinglimitsmathematical}, self-training \cite{ranaldi2025multilingual}, and code generation \cite{chen2025predicate}. In this research, we expand the use of GRPO to hateful meme analysis by optimizing thinking-based MLLMs and studying how CoT supervision interacts with RL-based post-training for improved detection and explanation.}

%GRPO has been effectively applied across a range of domains, including mathematical reasoning \cite{shao2024deepseekmathpushinglimitsmathematical, ding2025multi, chen2025spectral, nan2025ngrpo, tan2025gtpo, ren-2025-lsrl}, self-training \cite{wei2025sftsecondrlupt, ranaldi2025multilingual}, and code generation \cite{chen2025predicate, fan2025posterior, pennino2025reasoning, robeyns2025improving}. However, its application to hateful meme analysis remains largely unexplored, with only very recent or concurrent work—most notably \cite{mei2025expo}—investigating the use of RL and GRPO in this setting. In this work, we extend this line of research by optimizing thinking-based MLLMs with GRPO and systematically studying the role of chain-of-thought supervision for hateful meme detection.
%
%
%

\section{Datasets}

\subsection{Dataset Preparation}
\label{sec:dataset_prepare}
For \textit{hateful meme} analysis, most prior work has primarily focused on coarse-grained classification \cite{cao2023pro}. 
Recent studies have shown that training models with more detailed supervision can significantly enhance task performance, as such information 
mimics human-like, step-by-step reasoning when analyzing multimodal inputs and making final decisions \cite{zhu2024distilling}. 
Effective training of thinking-based MLLMs requires providing sufficient and structured task-specific information beyond binary labels. 
Therefore, we curated existing datasets containing both binary and fine-grained labels and extended the original Hateful Memes dataset with explanations and rationales for the assigned labels, as well as step-by-step CoT reasoning traces.

\subsubsection{Hateful Memes -- Original Dataset}
\textit{Hateful Memes} \cite{kiela2020hateful} is one of the most widely used early benchmarks for multimodal hate speech detection, consisting of approximately 12k memes that combine textual and visual content. 
It was carefully curated to ensure that effective classification requires joint multimodal understanding over both modalities, as neither text nor image alone is sufficient to reliably determine the label. 
The dataset includes a mixture of synthetically generated memes and real-world examples and maintains a balanced distribution between hateful and non-hateful content.\footnote{In this work, we focus on the \textit{unseen} splits, as these were augmented with explanations in \cite{kmainasi2025memeintel}.} 
Table~\ref{tab:data_stat_hateful_meme} summarizes the dataset distribution.

\begin{table}[h]
\centering
\setlength{\tabcolsep}{3pt}
\scalebox{0.95}{%
\begin{tabular}{@{}lrrrr@{}}
\toprule
\multicolumn{1}{c}{\textbf{Class Label}} &
\multicolumn{1}{c}{\textbf{Train}} &
\multicolumn{1}{c}{\textbf{Dev}} &
\multicolumn{1}{c}{\textbf{Test}} &
\multicolumn{1}{c}{\textbf{Total}} \\ \midrule
Non-hateful & 5,481 & 340 & 1,250 & 7,071 \\
Hateful     & 3,019 & 200 &   750 & 3,969 \\ \midrule
\textbf{Total} & \textbf{8,500} & \textbf{540} & \textbf{2,000} & \textbf{11,040} \\ \bottomrule
\end{tabular}
}
\caption{Distribution of labels in the dataset.}
\label{tab:data_stat_hateful_meme}
\end{table}

\subsubsection{Explanation/Reasoning}
We utilize the rationales introduced in the recently released MemeIntel work \cite{kmainasi2025memeintel}, which provides human-verified explanations corresponding to the binary labels.

\subsubsection{Fine-Grained Labels} %FHM
To obtain fine-grained annotations for the original  dataset, we leverage publicly available annotations provided by Mathias et al. \cite{mathias2021findings}, which categorize each hateful instance according to protected categories (religion, race, sex, disability, and nationality) and attack types (dehumanizing, inferiority, inciting violence, mocking, contempt, slurs, and exclusion). 
%
%
%human-annotated 
%
%
\subsection{Step-by-Step Reasoning -- Our Extension}
\label{ssec_step_by_step_reasoning}
The final extension for the dataset introduces step-by-step reasoning rationales, which are essential for training thinking-based MLLMs. However, collecting high-quality reasoning traces from human annotators is a highly time-consuming and labor-intensive process.
% Collecting such detailed reasoning traces directly from human annotators is a highly time-consuming and labor-intensive process. 
Meanwhile, recent advances in commercial reasoning-oriented models have demonstrated strong and reliable reasoning capabilities in practice~\cite{latif2025openaiO1,comanici2025gemini}. Motivated by these observations, we adopt a distillation-based approach to generate step-by-step reasoning traces, referred to as the chain-of-thought distillation (CoTD).
%annotations 
%

We use GPT-4.1 to generate intermediate step-by-step reasoning traces conditioned on the meme image, extracted text, annotation guidelines, and both binary and fine-grained labels. These private reasoning sequences are used exclusively to supervise MLLMs during training and are not exposed during inference. Our approach follows prior work on CoTD for improving model reasoning performance \cite{zhu2024distilling}. To prevent label leakage, the model is explicitly prompted to reason independently, without copying or paraphrasing the reference explanations. The prompting templates are released in the GitHub repository.

% The complete prompting setup for generating hatefulness-focused CoT reasoning is provided in 
% %Appendix~\ref{app:Chain-of-thoughs_generation}.
% Appendix~\ref{sec:synthetic_data_generation}.
% % \firoj{should we use llm-judge for an evaluation of the reasoning? we are claiming this is a key contribution}
%
%
%

\paragraph{CoT Evaluation with LLM-as-a-Judge.}
We evaluate the quality of the CoTD using an LLM-as-a-judge protocol, following the work of Kmainasi et al. \cite{kmainasi2025memeintel}. We employ strong vision--language models as automated judges to enable scalable and consistent evaluation. Each sample is assessed along four dimensions: \emph{Informativeness}, measuring the use of salient visual and textual evidence; \emph{Clarity}, assessing logical structure and ease of interpretation; \emph{Plausibility}, evaluating whether the reasoning provides a coherent interpretation aligned with human judgment; and \emph{Faithfulness}, measuring grounding in observable meme content without hallucinated or unsupported claims.

We use two strong vision--language models, \textbf{InternVL3.5} and \textbf{Phi-3.5}, as independent judges, scoring each dimension on a 5-point Likert scale. Table~\ref{tab:llm_judge_compact} reports the resulting scores together with inter-judge agreement. Agreement on ordinal ratings is quantified using the $r^*_{wg(j)}$ index \cite{james1984estimating}, which contrasts the observed rating variance with the maximum variance under complete disagreement:
\[
r^*_{wg(j)} = 1 - \frac{S_X^2}{\sigma^2_{\text{mv}}},
\]
where $S_X^2$ denotes the variance between judges and $\sigma^2_{\text{mv}}$ is the maximum possible variance for a 5-point scale. Higher values indicate a stronger consensus. As shown in Table~\ref{tab:llm_judge_compact}, agreement scores are consistently high across all dimensions, indicating reliable and stable LLM-based judgments. The complete evaluation assets are released in the supplementary material.

Both judges assign average scores above 4.0, indicating a strong overall explanation quality. \emph{Faithfulness} and \emph{Clarity} receive the highest scores from both InternVL3.5 and Phi-3.5, suggesting that the generated reasoning traces are well grounded in observable meme content and are logically structured and easy to follow.

\begin{table}[t]
\centering
\small
\setlength{\tabcolsep}{4pt}
\begin{tabular}{lccc}
\toprule
\textbf{Metric} & \textbf{InternVL3.5} & \textbf{Phi-3.5} & \textbf{Agreement} \\
\midrule
Faithfulness     & \textbf{4.782} & 4.557 & 0.944 \\
Clarity          & \textbf{4.733} & 4.537 & 0.943 \\
Plausibility     & \textbf{4.608} & 4.241 & 0.934 \\
Informativeness  & \textbf{4.399} & 4.097 & 0.942 \\
\midrule
Average          & \textbf{4.631} & 4.358 & 0.940 \\
\bottomrule
\end{tabular}
\caption{\textbf{Train+Dev averaged LLM-as-a-judge scores (1--5) and agreement indices.}
\textbf{InternVL3.5} = OpenGVLab InternVL3.5-8B, \textbf{Phi-3.5} = Phi-3.5-Vision-Instruct. \textbf{Agreement} denotes inter-judge agreement.}
\label{tab:llm_judge_compact}
\end{table}

\section{Methodology}
\label{sec_methodology}
In this section, we present our proposed RL-based method for hateful meme analysis using thinking-based MLLMs, which is shown in Figure \ref{fig:overview}. Our approach is designed to jointly train MLLMs to perform classification, explanation generation, and step-by-step reasoning through a structured output format and a multi-stage training process.

\paragraph{Framework Overview.}
As illustrated in Figure~\ref{fig:overview}, we propose an end-to-end hateful meme analysis framework that integrates fine-grained supervision, CoT reasoning, and RL-based post-training. Given a raw meme consisting of an image and embedded text, we first construct instruction-following training examples by combining multiple supervision signals, including a binary hateful/non-hateful label, fine-grained annotations specifying the protected category and attack type, OCR-extracted text, and classification guidelines. This unified representation encourages the model to jointly reason over visual and textual cues while grounding its predictions in established hate taxonomies.

To enable step-by-step reasoning, we adopt a weakly supervised CoT distillation strategy, in which a strong commercial MLLM is used to generate private reasoning traces conditioned on the meme content, labels, and guidelines. These distilled CoT rationales are used exclusively during training. Model optimization follows a two-stage post-training pipeline. First, supervised SFT aligns the model with the structured output format, gold labels, explanations, and distilled CoT supervision. Second, GRPO further refines the model by sampling multiple candidates (private CoT + label + explanation/rationale) per meme and reinforcing those that achieve higher relative rewards, jointly optimizing label correctness, format compliance, and explanation quality. During inference, the model performs internal reasoning and outputs a predicted label together with a supporting rationale.

%
% We first formalize the task and output representations, then describe a two-stage optimization procedure consisting of supervised fine-tuning (SFT) followed by Group Relative Policy Optimization (GRPO) with a structured reward. Finally, we summarize the overall training pipeline.
%
In the following subsections, we first formalize the task.
% and briefly present the dataset preparation. 
We then describe the MLLM optimization procedure in multiple steps. Finally, we provide the detailed setup for training and evaluation.

\subsection{Task Formulation}
\textcolor{black}{
Given a meme $x$ consisting of an image and extracted text, the task is to predict whether the meme is hateful and to generate a natural language rationale that justifies the decision. 
Formally, given an instruction-style prompt $c(x)$, the model produces a structured output $y$ of the form:
}
%\todo{It sounds weird to me that our objective is to train a model. Isn't the task detecting hateful memes with explanations?   }
\[
y =
\big\langle
\texttt{<think>}~t~\texttt{</think>}~
\texttt{Label:}~\hat{\ell}~ 
\texttt{Explanation:}~\hat{e}
\big\rangle,
\]
where $t$ denotes a private CoT reasoning trace, $\hat{\ell}$ represents the predicted class label, and $\hat{e}$ is a natural language explanation that justifies the predicted label.
Let ${D}$ represent the labeled dataset as:
\[
\mathcal{D}
=
\{(x_i, y_i^\star)\}_{i=1}^{N},
\]
where $y_i^\star$ contains the gold label, explanation, and distilled teacher reasoning. $i$ denotes sample number and $N$ is the total number of samples in ${D}$.

\subsection{Training \textit{thinking-based MLLMs}}
\label{ssec:train_llm_procedure}
%
%After preparing the dataset ${D}$, we train the MLLM with it.
%
Our training procedure consists of two steps: first initializing the MLLM parameters via SFT and next applying GRPO. Below we provide brief details of the training procedure.
\subsubsection{SFT warm-up}
We initialize the MLLM via SFT on $\mathcal{D}$. Let $\pi_\theta$ denote the conditional autoregressive model parameterized by $\theta$. Given an instruction-style prompt $c(x)$ that encodes the meme image and OCR-extracted text, the model is trained to generate the structured target output $y^\star$.
Parameters $\theta$ are optimized by minimizing the negative log-likelihood loss:
\begin{equation}
\mathcal{L}_{\text{SFT}}(\theta)
=
\mathbb{E}_{(c,y^\star)\sim\mathcal{D}}
\left[
-\log\pi_{\theta}(y^\star\mid c)
\right]
\end{equation}
$\mathbb{E}_{(c,y^\star)\sim\mathcal{D}}$ denotes the empirical expectation over the training set (approximated by minibatch averages), i.e., we average the negative log-likelihood of the target output $y^\star$ conditioned on the prompt $c$.
This stage aligns the model with gold labels, explanations, and distilled reasoning traces, providing a strong initialization for subsequent GRPO-based optimization.

After the MLLM is initialized, we select the best checkpoint (for the next step) based on the evaluation loss on a held-out validation set.
%
%
%

%\subsubsection{Stage II: Group Relative Policy Optimization (GRPO) for Reasoning and Explanation Quality}
\subsubsection{GRPO Optimization}
\label{subsec:stage2-grpo}

%GRPO is a reinforcement learning objective designed to improve reasoning quality by comparing multiple candidate outputs generated for the same input and optimizing the model based on their relative performance. Instead of relying on an external reward model, GRPO uses group-wise normalization to encourage better reasoning trajectories while maintaining stable policy updates.
GRPO optimizes the model by contrasting multiple sampled outputs per input and reinforcing reasoning trajectories that outperform their peers, enabling stable and effective reasoning-centric post-training. It optimizes the MLLM to improve reasoning and explanation quality. 

In this stage, the MLLM is treated as a stochastic policy $\pi_\theta$, which generates outputs conditioned on the prompt $c(x)$. For each input, we sample a group of $K$ candidate outputs %$\{y_{1},\dots,y_{K}\}$ 
$\{y_{1},\dots,y_{K}\}$
from $\pi_\theta$ and compute a scalar reward for each completion.
The overall reward function is defined as a weighted combination of multiple components:
\begin{align}
R(y)
&=
\alpha_{\text{fmt}} R_{\text{fmt}}(y)
+ \alpha_{\text{lbl}} R_{\text{lbl}}(y) \nonumber\\
&\quad
+ \alpha_{\text{len}} R_{\text{len}}(y)
+ \alpha_{\text{met}} R_{\text{met}}(y),
\end{align}
with weights $\alpha_{\text{fmt}}=0.5$, $\alpha_{\text{lbl}}=0.4$, $\alpha_{\text{len}}=0.05$ and $\alpha_{\text{met}}=0.05$, which are set empirically. The individual components of $R(y)$ are:
%$R_{\text{fmt}}$ enforces adherence to the required output format, $R_{\text{lbl}}$ rewards correct label prediction, $R_{\text{len}}$ softly regularizes explanation length toward the target, and $R_{\text{met}}$ encourages semantic similarity between generated and reference explanations.
\begin{itemize} %[noitemsep,topsep=0pt,labelsep=.5em]
%[leftmargin=1.2em]
    \item $\mathbf{R_{\text{fmt}}(y)}$ \textbf{(format)}: encourages structurally consistent outputs with explicit reasoning, prediction, and explanation components.
    \item $\mathbf{R_{\text{lbl}}(y)}$ (\textbf{label correctness}): rewards selecting the correct class compared to the gold annotation, ensuring the reasoning is anchored to the right decision.
    \item $\mathbf{R_{\text{len}}(y)}$ (\textbf{length regularization}): softly encourages rationales to be near the target length (about 100 words, matching \cite{kmainasi2025memeintel} gold rationales), discouraging overly short or excessively verbose explanations: $R_{\text{len}}(y)=\exp\!\left(-\frac{(L-100)^2}{2\sigma^2}\right)$. \textcolor{black}{We choose $\sigma=20$ so that the length reward acts as a soft regularizer, gently discouraging extreme deviations from the target length without dominating the GRPO objective.}

%\todo{Mohamed, any justification for setting it to 20? }

    % \[
    % R_{\text{len}}(y)
    % = \exp\!\left(-\frac{(L-100)^2}{2\sigma^2}\right),
    % \qquad
    % \sigma=20.
    % \]
    %
    \item $\mathbf{R_{\text{met}}(y)}$ (\textbf{semantic similarity}): rewards explanations that are semantically close to the gold rationale via METEOR score~\cite{banerjee2005meteor},\footnote{We choose METEOR because it can be computed efficiently during training.} encouraging higher-quality justifications beyond label accuracy. 
\end{itemize}

\paragraph{Objective function.}
GRPO uses the group-average reward $\bar{R}$ as a baseline to compute normalized advantages:
\vspace{-0.15cm}
\[
\bar{R}=\frac{1}{K}\sum_{k=1}^{K} R_k,
\qquad
A_k = \frac{R_k-\bar{R}}{\mathrm{std}(\{R_j\})+\varepsilon}
\]
Let $\pi_{\theta_{\text{old}}}$ denote the policy before the update. For each generated token $a_{k,t}$ at state $h_{k,t}$, the importance ratio is as follows. 
\vspace{-0.2cm}
\[
r_{k,t}
=
\frac{\pi_\theta(a_{k,t}\mid h_{k,t})}
     {\pi_{\theta_{\text{old}}}(a_{k,t}\mid h_{k,t})}
\]
\vspace{-0.1cm}
The GRPO objective maximizes the clipped surrogate with KL regularization:
\begin{align}
\mathcal{J}_{\text{G}}(\theta)
&=
\mathbb{E}\Bigg[
\sum_{k=1}^{K}\sum_{t}
\begin{aligned}[t]
\min\!\Big(
& r_{k,t}A_k,\\
& \mathrm{clip}(r_{k,t},1-\epsilon,1+\epsilon)A_k
\Big)
\end{aligned}
\nonumber\\
&\qquad
-\beta\,
\mathrm{KL}\!\left(
\pi_\theta(\cdot\mid h_{k,t})
\;\middle\|\;
\pi_{\text{ref}}(\cdot\mid h_{k,t})
\right)
\Bigg]
\end{align}
%\todo{Mohamed, There are many parameters and some of them are not defined. For instance, beta parameter is not defined. Even though they are parameters of KL regularization. It is better to briefly mention that.}

\noindent
\textcolor{black}{
where $\epsilon$ denotes the PPO-style clipping threshold, $\beta$ controls the contribution of the KL regularization term, and $\pi_{\text{ref}}$ is a fixed reference policy initialized from the SFT checkpoint. The KL term constrains the updated policy to remain close to the reference model, improving training stability and preventing reward over-optimization.
}

Overall, this optimization encourages higher-reward reasoning trajectories while stabilizing updates via clipping and KL regularization. %\todo{Mohamed, I remember that you said we are using a different reward function than prior work. We might emphasize it. The current text suggests it is what GRPO does originally. } %\textcolor{red}{I emphasized the reward function that jointly optimizes the explanation and classification. The other reward function was for unsupervised learning, where we kept it for the journal extension}

\section{Experimental Setup}
% Our experimental procedure begins by selecting a suitable \textit{thinking-based MLLM} to be fine-tuned on the dataset $\mathcal{D}$ prepared in Section~\ref{sec:dataset_prepare}. To this end, we first conduct experiments to identify the most appropriate base \textit{thinking-based MLLM}, as discussed in Section~\ref{ssec:_base_model_selection}. Next, we fine-tune this model following the procedure outlined in Section~\ref{ssec:train_llm_procedure}. Finally, we evaluate the resulting model, compare it against state-of-the-art approaches, and perform detailed analyses of model configurations in Sections~\ref{ssec:sota_comparison} and~\ref{ssec:ablation_study}.

\subsection{Models}
\label{ssec:_base_model_selection}
% \paragraph{Multimodal classification baseline.}
% \todo[inline]{ why Qwen is selected? may be citing any paper}

For our experiments, we use both open- and closed-weight multimodal LLMs. Our open-weight baselines include Llama-3.2-11B, Llama-4-Scout-17B~\cite{dubey2024llama}, Qwen3-VL-8B~\cite{li2026qwen3} and Gemma-3-12B~\cite{team2025gemma}.
% and Kimi-VL-A3B~\cite{team2025kimi}. 
As a representative closed-weight model, we use GPT-4.1~\cite{openai2023gpt}. For CoT experiments, we use Qwen3-VL-8B-Thinking~\cite{bai2025qwen3vltechnicalreport}, a reasoning-enhanced checkpoint released with the Qwen3-VL family, which enables controlled CoT ablations in an open-weight setting and is reported in \cite{li2026qwen3} to perform strongly on multimodal reasoning benchmarks such as MMMU \cite{yue2024mmmu} and MathVista \cite{lu2024mathvista}. Overall, we select these models because their technical reports and model cards report strong results on established vision--language benchmarks, making them competitive baselines for multimodal understanding tasks. Our experiments include both thinking and instruction-tuned models.

\subsection{Zero-shot and CoT Inference}
\label{ssec:zero_shot_cot}
We benchmark the open- and closed-weight MLLMs described above, including both instruction-tuned and explicit ``thinking'' variants, under zero-shot and CoT prompting. For fairness, we disable the internal thinking mode for thinking-enabled models in non-CoT settings. \textcolor{black}{This setup assesses the reasoning capabilities of pretrained MLLMs without task-specific fine-tuning, serving as a reference point for understanding how much performance can be achieved from general-purpose multimodal reasoning alone.}

%\textcolor{red}{estimates \todo{Mohamed, the word "estimate" does not seem right here.} an upper bound on reasoning performance without task-specific training. \todo{Mohamed, not clear to me why it is upper bound.}
%}

% We benchmark both open-source and closed-source multimodal large language models (MLLMs), including instruction-tuned and explicit ``thinking'' variants, under zero-shot and chain-of-thought (CoT) prompting settings. To ensure a fair comparison, the internal thinking mechanism of thinking-enabled models is disabled in non-CoT experiments. This evaluation setting is designed to assess the upper bounds of reasoning performance without any task-specific training. 

\subsection{SFT warm-up}
\label{ssec_sft}
Prior work has demonstrated that performing SFT before RL-based optimization improves both training stability and downstream performance in \textcolor{black}{reasoning-oriented LLMs \cite{shao2024deepseekmathpushinglimitsmathematical, 
wei2025sftsecondrlupt}}. %\todo{Mohamed, we need a reference for that prior work} 
Motivated by these findings, we introduce an SFT warm-up stage prior to GRPO to inject task-specific knowledge and to bootstrap the structured reasoning format required for fine-grained hateful meme analysis. 

% All SFT checkpoints are evaluated both as standalone models and as initialization points for subsequent GRPO training. \todo{I am a bit confused with this paragraph. Is it different that Section 4.2.1 ? If not, then this paragraph seems unnecessary. It does not provide info about the experiments.}

\paragraph{SFT without explicit CoT}
We consider two SFT variants that do not include explicit CoT supervision. The first variant is trained on a mixture of binary classification labels (hateful vs.\ non-hateful), fine-grained annotations (protected category and attack type), and human-written explanations, denoted \textbf{SFT warm-up  (Cls + FG + Exp, No-CoT)}. The second variant excludes fine-grained labels and is trained only on binary labels and explanations, denoted \textbf{SFT warm-up  (Cls + Exp, No-CoT)}.  
In both settings, the model is initialized from a pretrained thinking-based backbone. To preserve its reasoning capability and maintain output-format consistency, empty \texttt{<think></think>} tags are included as a prefix to the assistant output during training and excluding them from loss calculation.

\paragraph{SFT with distilled CoT}
To study the effect of explicit reasoning supervision during SFT, we further fine-tune the \textbf{SFT warm-up  (Cls + FG + Exp, No-CoT)} checkpoint using GPT-4.1 generated step-by-step reasoning traces placed inside the \texttt{<think>...</think>} block (See section \ref{ssec_step_by_step_reasoning}). This variant, denoted \textbf{SFT warm-up  (Cls + FG + Exp, CoTD)}, exposes the model to fine-grained reasoning signals during SFT while preserving the same output structure.
Although explicit CoT supervision is absent in some SFT variants, the model can still generate internal reasoning at inference time due to its pretrained thinking-based initialization. Overall, the SFT warm-up stage primarily aligns the model with task supervision and output structure, while GRPO further refines reasoning and explanation quality.

\subsubsection{Supervised GRPO}

We apply supervised GRPO using multiple initialization checkpoints to analyze the effect of different SFT warm-up strategies. As a baseline, we initialize GRPO directly from the pretrained thinking-based backbone, denoted \textbf{Qwen3-VL-8B-Thinking + GRPO}. In addition, we initialize GRPO from each SFT warm-up variant: \textbf{SFT warm-up  (Cls + FG + Exp, No-CoT)}, \textbf{SFT warm-up  (Cls + Exp, No-CoT)}, and \textbf{SFT warm-up  (Cls + FG + Exp, Distilled CoT)}. All models are optimized using supervised GRPO on the labeled hateful meme dataset.
As described in Section~\ref{subsec:stage2-grpo}, GRPO samples $K$ candidate completions per input and updates the policy using a composite reward function that promotes structurally valid outputs, correct predictions, and high-quality explanations, while supporting stable optimization.

% As described in Section~\ref{subsec:stage2-grpo}, GRPO samples $K$ candidate completions per input and updates the policy based on a composite reward function, encourages structurally valid outputs, correct decisions, and high-quality explanations while maintaining stable optimization. 
% that jointly accounts for output format compliance, \textit{label correctness, explanation length regularization, and semantic similarity to gold rationales}. The reward is defined as:
% % \label{subsec:stage2-grpo}
% \[
% R(y) = \sum_{k \in \{\text{fmt},\,\text{lbl},\,\text{len},\,\text{met}\}}
% \alpha_k\, R_k(y)
% \]
% with weights $(0.5, 0.4, 0.05, 0.05)$, respectively. 
% This formulation encourages structurally valid outputs, correct decisions, and high-quality explanations while maintaining stable optimization.

\subsection{Training Setup}
All models are trained using full-parameter optimization with DeepSpeed ZeRO-3 on 4 NVIDIA H200 GPUs. SFT is performed for 3 epochs, with the best checkpoint selected based on development-set loss and evaluation conducted after each epoch. The \textbf{SFT (Cls + FG + Exp, Distilled CoT)}\footnote{Cls: class labels; FG: fine-grained labels; Exp: explanation.} model is initialized from the corresponding \textbf{SFT (Cls + FG + Exp, No-CoT)} checkpoint and further fine-tuned for an additional 3 epochs. For SFT, the per-device batch size is set to 4 with gradient accumulation steps fixed to 1. Optimization is carried out using AdamW with $\beta_1=0.9$, $\beta_2=0.95$, $\epsilon=1\times10^{-8}$, weight decay $0.1$, and gradient clipping at $1.0$. A cosine learning-rate scheduler with a Warm-up  ratio of $0.05$ is used throughout, and model selection is based on development loss.

GRPO is initialized from the corresponding SFT checkpoints and uses the same optimization, scheduler, and infrastructure settings, except that the per-device batch size is reduced to 2 due to increased memory demands. GRPO-specific hyperparameters include a KL-divergence coefficient $\beta=0.04$, PPO clip range $0.2$, value loss coefficient $0.1$, and GAE parameter $\lambda=0.95$. During policy optimization, \textcolor{black}{we sample 16 generations per input} with a maximum completion length of 4096 tokens, temperature 1.0, and top-$p$ sampling set to 0.85. All experiments are run with a fixed random seed of 42. %\todo{what is the value of K (candidates for GRPO)? } 

\subsection{Evaluation Setup and Metrics}
We follow the evaluation protocol of \cite{kmainasi2025memeintel}. For classification, we report Accuracy \textit{(Acc)}, Weighted F1 \textit{(W-F1)}, and Macro F1 \textit{(M-F1)}. Explanation quality is evaluated using BERTScore \textit{(BS)} \cite{zhang2020bertscoreevaluatingtextgeneration} and METEOR \textit{(MET)} \cite{banerjee-lavie-2005-meteor}.
%\todo{Mohamed, we can give reference to BERTScore and METEOR }

\section{Results and Analysis}

\subsection{Comparison with State-of-the-Art}
\label{ssec:sota_comparison}

Table~\ref{tab:sota_comparison} presents the quantitative comparison, showing that our proposed approach achieves the strongest accuracy and comparable explanation quality\footnote{All baseline results are reported as stated in the original papers and are not re-implemented or re-evaluated.}. 

% \todo{Mohamed, we need to explain how we got the results for the other studies. If they are the reported ones,  we should explicitly mention it.} 

Specifically, it attains 81.2\% Accuracy, 0.81 Weighted F1, and 0.79 Macro F1 for classification. \textcolor{black}{ In addition, our model achieves a BERTScore of 0.78 and a METEOR score of 0.52, which are comparable to or higher than the strongest baseline (i.e., \cite{kmainasi2025memeintel}).} These results highlight the effectiveness of the proposed \textit{thinking-based MLLM} and the \textit{GRPO-based optimization strategy} for both classification and explanation-generation tasks. 
%\todo{Mohamed, since we compare only accuracy,  we should not make bold statements about other metrics. Is there a way to say that 0.78 BertScore is good enough? We need to discuss what 0.78 BS and 0.49 MET suggest.}
% \todo[inline]{Please add more SOTA comparison here.}
% wu2024multimodal
\begin{table}[h]
\centering
\setlength{\tabcolsep}{2pt}
\scalebox{1.0}{
\begin{tabular}{lccccc} 
\toprule
\textbf{Model} & \textbf{Acc.} & \textbf{W-F1} & \textbf{M-F1} & \textbf{BS} & \textbf{MET} \\
\bottomrule

\cite{kiela2020hateful}                 & 69.47±2.06 & --   & --   & --   & --   \\
\cite{cao-etal-2022-prompting}                 & 72.98±1.09 & --   & --   & --   & --   \\
\cite{wu2024multimodal}                 & 76.4 & --   & --   & --   & --   \\
\cite{yang2024uncertainty}                 & 77.2 & --   & --   & --   & --   \\

\cite{burbi2023mapping}                 & 77.7 & --   & --   & --   & --   \\

\cite{mei2024improving}                 & 78.8 & --   & --   & --   & --   \\

\cite{kmainasi2025memeintel} & 79.9 & 0.80 & 0.79 & 0.78 & 0.49 \\ \midrule
\textbf{Proposed}            & \textbf{81.2} & \textbf{0.81} & \textbf{0.79} & \textbf{0.78} & \textbf{0.52} \\
\bottomrule
\end{tabular}
}
\caption{Comparison with SOTA and our results.}
\label{tab:sota_comparison}
\end{table}

\subsubsection{Zero-shot and CoT Inference}
%
% We evaluate both open-source and closed-source multimodal large language models (MLLMs), including instruction-tuned and explicit ``thinking'' variants, under zero-shot and chain-of-thought (CoT) prompting settings. To ensure a fair comparison, the internal thinking mechanism of thinking-enabled models is disabled in non-CoT experiments. This evaluation setting is designed to assess the upper bounds of reasoning performance without any task-specific training. 
%

%Table~\ref{tab:pretrained_mllms_fhm_only} reports the results obtained using \textcolor{red}{explanation-based} prompting, measured in terms of both classification performance and explanation quality. \firoj{need some discussion about the results}

Table~\ref{tab:pretrained_mllms_fhm_only} reports the results of the experiments described in Section~\ref{ssec:zero_shot_cot}. It benchmarks both open- and closed-source thinking-based pretrained MLLMs using classification and explanation-generation metrics. The results show that \textit{no single model consistently achieves the highest performance across all metrics}. Specifically, L4-Scout attains the highest accuracy and BERTScore, the commercial model GPT-4.1 with CoT prompting achieves the best F1 scores, while Qwen perform best on the METEOR score. Notably, despite being state-of-the-art, neither non-commercial thinking-based models nor CoT-based approaches uniformly dominate across evaluation criteria. 

\begin{table}[t]
\centering
\setlength{\tabcolsep}{2pt}
\scalebox{1.0}{%
\begin{tabular}{lccccc}
\toprule
\textbf{Model} & \textbf{Acc.} & \textbf{W-F1} & \textbf{M-F1} & \textbf{BS} & \textbf{Met.} \\
\midrule
Llama +CoT   & 60.0 & 0.606 & 0.592 & 0.600 & 0.165 \\
Llama        & 64.2 & 0.599 & 0.547 & 0.660 & 0.234 \\
$\Delta$     & -4.2 & +0.007 & +0.045 & -0.060 & -0.069 \\
\midrule
Qwen-I +CoT  & 65.3 & 0.658 & 0.652 & 0.646 & \textbf{0.263} \\
Qwen-I       & 63.0 & 0.630 & 0.630 & 0.662 & 0.216 \\
$\Delta$     & +2.3 & +0.028 & +0.022 & -0.016 & +0.047 \\
\midrule
Qwen-T +CoT  & 70.2 & 0.706 & 0.694 & 0.581 & 0.202 \\
Qwen-T       & 65.6 & 0.661 & 0.652 & 0.659 & 0.215 \\
$\Delta$     & +4.6 & +0.045 & +0.042 & -0.078 & -0.013 \\
\midrule
Gemma +CoT   & 65.5 & 0.659 & 0.655 & 0.621 & 0.135 \\
Gemma        & 65.2 & 0.654 & 0.652 & 0.664 & 0.215 \\
$\Delta$     & +0.3 & +0.005 & +0.003 & -0.043 & -0.080 \\
\midrule
L4-Scout +CoT& 75.4 & 0.687 & 0.545 & 0.596 & 0.111 \\
L4-Scout     & \textbf{77.0} & 0.710 & 0.580 & \textbf{0.653} & 0.252 \\
$\Delta$     & -1.6 & -0.023 & -0.035 & -0.057 & -0.141 \\
\midrule
GPT-4.1 +CoT & 74.1 & \textbf{0.739} & \textbf{0.719} & 0.470 & 0.072 \\
GPT-4.1      & 71.3 & 0.715 & 0.700 & 0.595 & 0.159 \\
$\Delta$     & +2.8 & +0.024 & +0.019 & -0.125 & -0.087 \\
%\midrule
%Kimi-I +CoT  & 0.677 & 0.667 & 0.636 & 0.621 & 0.186 \\
%Kimi-I       & 0.670 & 0.625 & 0.573 & \textbf{0.678} & 0.257 \\
%$\Delta$     & +0.007 & +0.042 & +0.063 & -0.057 & -0.071 \\
%\midrule
%Kimi-T +CoT  & 0.705 & 0.704 & 0.685 & 0.614 & 0.155 \\
%Kimi-T       & 0.675 & 0.668 & 0.639 & 0.660 & 0.222 \\
%$\Delta$     & +0.030 & +0.036 & +0.046 & -0.046 & -0.067 \\
\bottomrule
\end{tabular}
}
\caption{\textbf{Performance of pretrained MLLMs.}
Llama = Llama-3.2-11B; L4-Scout = Llama-4-Scout-17B; Qwen-I/T = Qwen3-VL-8B (Instruct/Thinking);
Gemma = Gemma-3-12B.
$\Delta$ denotes (CoT $-$ no-CoT).
Acc.: Accuracy; W-F1: weighted F1; M-F1: macro F1; BS: BERTScore; Met.: METEOR.}
\label{tab:pretrained_mllms_fhm_only}
\end{table}

% \todo{Mohamed, we are not discussing the impact of CoT. For instance, in all cases it decreases BS. In most of the cases, it increases W-F1.}

\paragraph{Impact of CoT prompting.}
Across all evaluated models, CoT prompting exhibits a \emph{systematic and asymmetric effect} on classification versus explanation metrics. On the one hand, CoT improves classification performance for most models, yielding consistent gains in Weighted and Macro F1 (up to +4.5 points for Qwen-T). A finer-grained error analysis reveals that these improvements are primarily driven by a reduction in false negatives, indicating that CoT helps models better identify subtle or implicit hateful content that is often missed under direct prompting. On the other hand, CoT \emph{consistently degrades explanation quality} as measured by BERTScore, and in most cases also by METEOR. This degradation is observed across all architectures, including both open-source and commercial models, with BERTScore drops ranging from $-1.6$ to $-12.5$. This pattern suggests that while CoT enhances internal reasoning for decision making, it shifts the output distribution away from the concise, annotation-style explanations used as references.

\paragraph{Model-dependent CoT behavior.}
The magnitude and direction of CoT effects vary substantially across model families. Reasoning-oriented models such as Qwen-T benefit the most from CoT, achieving the largest accuracy gains, which suggests a favorable alignment between CoT prompting and their pretraining objectives. In contrast, Llama-based models show weaker or even negative accuracy changes, indicating that explicit CoT may interfere with their implicit reasoning mechanisms. Interestingly, large frontier models such as GPT-4.1 exhibit improved classification performance with CoT but suffer the largest drops in explanation quality, highlighting a pronounced trade-off between decision accuracy and explanation calibration.

\paragraph{Model selection}
\textcolor{black}{These observations highlight the need for further investigation into how to effectively exploit open-source thinking-based MLLMs that are trainable under realistic computational budgets and allow reproducible post-training experiments.} Based on overall performance and ease of implementation (possible to train on commonly available GPUs), we select the Qwen model as the base MLLM for our subsequent experiments.

\subsubsection{Analysis of RL Post-Training}

Table~\ref{tab:sft_grpo_analysis} summarizes the impact of SFT warm-up and supervised GRPO on both classification performance and explanation quality. SFT warm-up alone already yields consistent improvements over cold-start (no warm-up) GRPO. In particular, \textbf{SFT warm-up  (Cls + FG + Exp)} achieves 78.1\% accuracy and macro-F1 of 0.77, outperforming the variant trained without fine-grained supervision (77.0\% accuracy and 0.75 macro-F1). This result indicates that \textit{incorporating fine-grained labels improves decision consistency} even in the absence of explicit reasoning supervision.

Adding distilled CoT supervision during SFT leads to further gains. \textbf{SFT warm-up  (Cls + FG + Exp, CoTD)} attains the strongest warm-up performance, reaching 79.2\% accuracy and 0.78 macro-F1, while also improving explanation quality, with BERTScore increasing from 0.77 to 0.78 and METEOR from 0.48 to 0.50. These improvements suggest that \textit{explicit reasoning supervision during SFT primarily enhances explanation fidelity in addition to classification performance}.

When combined with supervised GRPO, \textit{all SFT-initialized models substantially outperform} both prior SFT-only models and the cold-start (no warm-up) GRPO-trained model. This result demonstrates that \textit{GRPO without proper warm-up or initialization yields limited benefits}, as the cold-start variant achieves only 76.8\% accuracy and 0.75 macro-F1, along with noticeably weaker explanation quality (BERTScore 0.73), underscoring the importance of SFT warm-up. In contrast, initializing GRPO from \textbf{SFT warm-up } consistently improves performance: both \textit{SFT-Cls+Exp $\rightarrow$ GRPO} and \textit{SFT-Cls+FG+Exp $\rightarrow$ GRPO} improves accuracy (from 77.0\% to 80.4\% and from 78.1\% to 81.1\%), macro-F1 (from 0.75 to 0.78 and from 0.77 to 0.79), and METEOR score (from 0.48 to 0.50 and from 0.48 to 0.52).

We next analyze the influence of distilled CoT supervision for GRPO. Consistent with our observations from SFT-only models, where the highest accuracy (79.2\%) is achieved when CoT supervision is used, CoT also yields the strongest and most consistent performance across all metrics for GRPO-based fine-tuned models. More specifically, \textit{models initialized from checkpoints trained with CoT supervision consistently achieve the best performance after GRPO optimization}, highlighting the complementary role of CoT distillation and reinforcement learning in enhancing both classification and explanation quality.

%The strongest overall performance is achieved by GRPO models initialized from fine-grained SFT warm-up, with or without distilled chain-of-thought supervision. Both \textbf{SFT-CE-FG $\rightarrow$ GRPO} and \textbf{SFT-CE-FG-CoT $\rightarrow$ GRPO} reach 0.81 accuracy and 0.81 weighted F1, surpassing the results reported in \cite{kmainasi2025memeintel}.  Notably, the CoT-initialized variant achieves the highest explanation quality, confirming that explicit reasoning alignment during warm-up facilitates more effective reinforcement learning of explanation generation.

\begin{table}[t]
\centering
\setlength{\tabcolsep}{2pt}
\scalebox{1.0}{%
\begin{tabular}{lccccc}
\toprule
\textbf{Model} & \textbf{Acc.} & \textbf{W-F1} & \textbf{M-F1} & \textbf{BS} & \textbf{Met.} \\
\midrule
\multicolumn{6}{c}{\textbf{SFT warm-up}} \\
\midrule
SFT (Cls+Exp)        & 77.0 & 0.77 & 0.75 & 0.77 & 0.48 \\
% , No-CoT
SFT (Cls+FG+Exp)     & 78.1 & 0.78 & 0.77 & 0.77 & 0.48 \\
% , No-CoT
SFT (Cls+FG+Exp, CoTD) & 79.2 & 0.79 & 0.78 & 0.78 & 0.50 \\
\midrule
\multicolumn{6}{c}{\textbf{RL / GRPO}} \\
\midrule
GRPO (Cold Start)            & 76.8 & 0.77 & 0.75 & 0.73 & 0.47 \\
SFT-Cls+Exp $\rightarrow$ GRPO    & 80.4 & 0.80 & 0.78 & 0.76 & 0.50 \\
SFT-Cls+FG+Exp $\rightarrow$ GRPO & \textbf{81.1} & \textbf{0.81} & 0.79 & 0.77 & \textbf{0.52} \\
SFT-Cls+FG+Exp-CoTD $\rightarrow$ GRPO
                             & \textbf{81.2} & \textbf{0.81} & \textbf{0.79} & \textbf{0.78} & \textbf{0.52} \\
\bottomrule
\end{tabular}%
}
% \vspace{-2mm}
\caption{\textbf{Effect of SFT warm-up and GRPO post-training on Hateful Memes.}
Cls = classification label; FG = fine-grained labels; Exp = explanation; CoTD =  chain-of-thought distillation}
% CE = cross-entropy.
\label{tab:sft_grpo_analysis}
\end{table}

% \begin{table*}[t]
% \centering
% \setlength{\tabcolsep}{6pt}
% \scalebox{0.92}{
% \begin{tabular}{lccccc}
% \toprule
% \textbf{Model} & \textbf{Acc} & \textbf{W-F1} & \textbf{M-F1} & \textbf{BS} & \textbf{METEOR} \\
% \midrule
% \multicolumn{6}{c}{\textbf{Multimodal SFT warm-up}} \\
% \midrule
% SFT-Warm-up  (Cls + FG + Exp, No-CoT)        & 0.78 & 0.78 & 0.77 & 0.77 & 0.48 \\
% SFT-Warm-up  (Cls + Exp, No-CoT)             & 0.77 & 0.77 & 0.75 & 0.77 & 0.48 \\
% SFT-Warm-up  (Cls + FG + Exp, Distilled CoT) & 0.79 & 0.79 & 0.78 & 0.78 & 0.50 \\
% \midrule
% \multicolumn{6}{c}{\textbf{Multimodal RL / GRPO}} \\
% \midrule
% %Prior SOTA (IEEE/CVF 2023)                  & 0.777 & --   & --   & --   & --   \\
% %Prior SOTA (EMNLP 2025, MemeIntel)          & 0.799 & 0.80 & 0.79 & 0.78 & 0.49 \\
% GRPO (Cold Start)                           & 0.77  & 0.77 & 0.75 & 0.73 & 0.47 \\
% SFT-CE-FG $\rightarrow$ GRPO                & \textbf{0.81} & \textbf{0.81} & 0.79 & 0.77 & \textbf{0.52} \\
% SFT-CE $\rightarrow$ GRPO                   & 0.80 & 0.80 & 0.78 & 0.76 & 0.50 \\
% SFT-CE-FG-CoT $\rightarrow$ GRPO            & \textbf{0.81} & \textbf{0.81} & \textbf{0.79} & \textbf{0.78} & \textbf{0.52} \\
% \bottomrule
% \end{tabular}
% }
% \caption{Effect of supervised fine-tuning (SFT) warm-up and supervised GRPO on the Hateful Memes benchmark. Metrics include accuracy (Acc), weighted F1 (W-F1), macro F1 (M-F1), BERTScore (BS), and METEOR (M).}
% \label{tab:sft_grpo_analysis}
% \end{table*}

\begin{figure}[t]
    \centering
    \includegraphics[width=\linewidth]{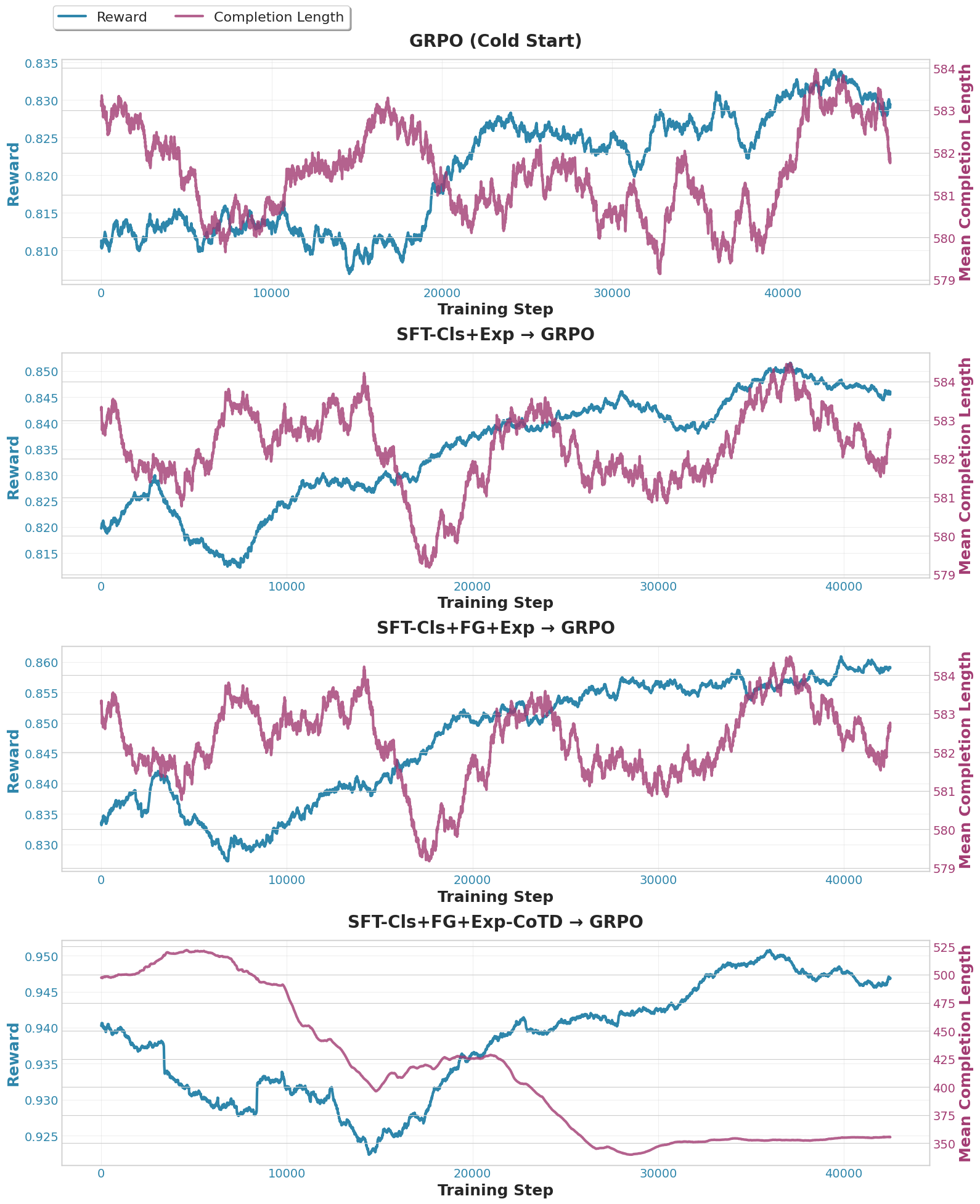}
    \caption{GRPO training dynamics under different initializations, showing mean group reward (left axis) and mean completion length (right axis). Moving average smoothing with window size of 1000 is applied.}
    \label{fig:sft_grpo_reward_length}
\end{figure}

\subsection{GRPO Training Dynamics under Different Initialization Regimes}
\label{sec:sft_grpo_dynamics}

Figure~\ref{fig:sft_grpo_reward_length} compares GRPO reward trajectories and mean completion length across initialization regimes (length is the average generated tokens per sample). Table~\ref{tab:sft_grpo_analysis} reports the corresponding end-task metrics. Overall, GRPO benefits strongly from SFT warm-up: warm-start runs show smoother reward increases and yield markedly better final accuracy and explanation quality than cold-start GRPO.

\textbf{Cold-start GRPO struggles without prior alignment.} When initialized directly from the pretrained backbone, GRPO exhibits noisier reward curves with only modest gains and largely stable completion length. This indicates that RL alone has difficulty simultaneously learning the required output structure and the task-specific decision boundary, which is reflected in the weakest BS/MET scores in Table~\ref{tab:sft_grpo_analysis}.

\textbf{SFT warm-up lets GRPO refine behavior instead of learning the format.} Initializing GRPO from SFT (Cls+Exp) yields large gains over SFT-only (e.g., 77.0$\rightarrow$80.4 Acc), suggesting that once the model is aligned to the target schema and supervision distribution, GRPO can focus on improving label correctness and rationale quality.

\textbf{Fine-grained supervision increases the ceiling.} Warm-up with protected-category and attack-type labels (Cls+FG+Exp) leads to stronger outcomes after GRPO (81.1 Acc, 0.52 Met), implying that fine-grained supervision improves decision consistency and supports better explanations beyond binary labels alone.

\textbf{CoTD warm-up improves metrics but reveals CoT collapse.} The CoTD-initialized run attains the best overall performance after GRPO (81.2 Acc, 0.78 BS, 0.52 MET), but Figure~\ref{fig:sft_grpo_reward_length} shows a pronounced drop in completion length during optimization. This suggests an RL shortcut where the policy boosts reward by compressing the \texttt{<think>} segment rather than consistently enriching reasoning, i.e., reward-driven \emph{CoT collapse}. In moderation settings, shorter generations may still preserve label reward while reducing verbosity-related risks.

\textbf{Implication.} These dynamics motivate reward designs that explicitly control the internal reasoning budget (e.g., adding $R_{\text{think-len}}$ or a target-range constraint) to discourage collapse and prevent the model from exploiting length shortcuts.

% \subsection{Qualitative Error Analysis}
% \label{ssec:error_analysis}

% To better understand the behavioral differences between SFT and GRPO-trained models, we conduct a qualitative error analysis, with representative cases presented in Appendix~\ref{app:error_analysis_app}. Overall, the analysis reveals complementary strengths and systematic weaknesses across training paradigms.

% GRPO substantially reduces false positives in benign or culturally grounded memes by explicitly verifying the absence of attack mechanisms, whereas SFT models tend to over-predict hate based on diffuse stereotype priors. However, this robustness comes at the cost of sensitivity to \emph{implicit symbolic hate}: GRPO models frequently misclassify historically charged or symbolically hateful content when explicit slurs or direct attacks are absent. In contrast, SFT models are more responsive to such symbolism but lack discrimination in ambiguous contexts.

% Across all training strategies, we observe consistent failures on memes requiring multi-hop or highly implicit reasoning, indicating a shared limitation in recovering indirect targets and harm pathways. Finally, models trained with fine-grained protected-category supervision exhibit clear advantages in difficult cases, as explicit target grounding enables correct interpretation where coarse-label models fail.

% Taken together, these findings suggest that GRPO and SFT capture complementary inductive biases, and that fine-grained supervision remains essential for robust detection of implicit hate.

\section{Conclusions}

We studied the extent to which thinking-based MLLMs can improve hateful meme understanding when explicitly optimized for both prediction and explanation quality. To this end, we introduced an RL-based post-training method with a GRPO objective that jointly rewards correct classification, structurally valid outputs, and semantically aligned explanations. We additionally constructed an extended dataset with weakly/pseudo-supervised distilled chain-of-thought rationales to provide fine-grained reasoning supervision. Across experiments on the Hateful Memes benchmark, our approach improves both classification performance and explanation quality, achieving state-of-the-art results and demonstrating the value of reinforcement post-training for multimodal hate analysis. Future work includes \textit{(i)} evaluating generalizability on additional hateful meme and multimodal toxicity benchmarks, including multilingual settings; \textit{(ii)} assessing transfer across a broader set of thinking-enabled MLLMs and prompting regimes; \textit{(iii)} extending the framework to weakly/pseudo-labeled training at scale to leverage large collections of unlabeled memes; and \textit{(iv)} exploring self-supervised GRPO objectives for unlabeled data. Beyond scaling and transfer, our training-dynamics analysis suggests several RL-specific directions: \textit{(v)} designing section-wise rewards that explicitly control the internal reasoning budget 
% (e.g., constraining `<think>` length separately from public rationales) 
to mitigate reward-driven thought collapse.

%Future work include \textit{(i)} evaluate the generalizability of our approach on additional hateful meme and multimodal toxicity datasets, including multilingual settings; \textit{(ii)} assess transfer across a broader set of thinking-enabled MLLMs and prompting regimes; \textit{(iii)} extend our framework to weakly/pseudo-labeled training at scale to leverage large collections of unlabeled memes; and \textit{(iv)} explore self-supervised GRPO objectives for unlabeled data. 

% Future studies should also incorporate stronger human-centered evaluation of explanation faithfulness and examine robustness under distribution shift, adversarial or coded hate, and evolving meme conventions.

% Future work:
% \begin{itemize}
%     \item Explore generalizability of the proposed approach for similar datasets.
%     \item Explore generalizability of the proposed approach for similar thinking models.
%     \item Extend the existing approach in the weakly/pesudo-labeled settings which will allow us to incorporate a wide collection of unlabeled data. 
%     \item Explore self-supervised training on unlabeled dataset with GRPO.    

% \end{itemize}

\section{Limitations}
Our results are based on the Hateful Memes benchmark and may not generalize across domains or languages. Weakly supervised CoT distillation from commercial MLLMs can introduce bias and unfaithful reasoning. GRPO increases computational cost and reliance on closed models may affect reproducibility. Explanation quality is assessed using automatic metrics that only partially reflect human judgment.

\section*{Acknowledgments}
The work of A. E. Shahroor and F. Alam was supported by NPRP grant 14C-0916-210015 from the Qatar National Research Fund, part of the Qatar Research Development and Innovation Council (QRDI). The findings reported herein are solely the responsibility of the authors.
%%
%% The next two lines define the bibliography style to be used, and
%% the bibliography file.
\bibliographystyle{ACM-Reference-Format}
\balance
\bibliography{bibliography/custom}

@inproceedings{lu2024mathvista,
  title     = {{MathVista}: Evaluating Mathematical Reasoning of Foundation Models in Visual Contexts},
  author    = {Lu, Pan and Bansal, Hritik and Xia, Tony and others},
  booktitle = {International Conference on Learning Representations (ICLR)},
  year      = {2024},
  url       = {https://arxiv.org/abs/2310.02255}
}

@inproceedings{yue2024mmmu,
  title     = {{MMMU}: A Massive Multi-discipline Multimodal Understanding and Reasoning Benchmark for Expert {AGI}},
  author    = {Yue, Xiang and Ni, Yuansheng and Zhang, Kai and others},
  booktitle = {Proceedings of the IEEE/CVF Conference on Computer Vision and Pattern Recognition (CVPR)},
  year      = {2024},
  url       = {https://arxiv.org/abs/2311.16502}
}

@article{openai2023gpt,
  title={GPT-4 technical report},
  author={OpenAI, R},
  journal={arXiv},
  pages={2303--08774},
  year={2023}
}

@article{dubey2024llama,
  title={The {Llama} 3 herd of models},
  author={Dubey, Abhimanyu and Jauhri, Abhinav and Pandey, Abhinav and Kadian, Abhishek and Al-Dahle, Ahmad and Letman, Aiesha and Mathur, Akhil and Schelten, Alan and Yang, Amy and Fan, Angela and others},
  journal={arXiv preprint arXiv:2407.21783},
  year={2024}
}

@article{team2025gemma,
  title={Gemma 3 technical report},
  author={Team, Gemma and Kamath, Aishwarya and Ferret, Johan and Pathak, Shreya and Vieillard, Nino and Merhej, Ramona and Perrin, Sarah and Matejovicova, Tatiana and Ram{\'e}, Alexandre and Rivi{\`e}re, Morgane and others},
  journal={arXiv preprint arXiv:2503.19786},
  year={2025}
}

@inproceedings{banerjee2005meteor,
  title={{METEOR}: An automatic metric for MT evaluation with improved correlation with human judgments},
  author={Banerjee, Satanjeev and Lavie, Alon},
  booktitle={Proceedings of the acl workshop on intrinsic and extrinsic evaluation measures for machine translation and/or summarization},
  pages={65--72},
  year={2005}
}

@article{comanici2025gemini,
  title={Gemini 2.5: Pushing the frontier with advanced reasoning, multimodality, long context, and next generation agentic capabilities},
  author={Comanici, Gheorghe and Bieber, Eric and Schaekermann, Mike and Pasupat, Ice and Sachdeva, Noveen and Dhillon, Inderjit and Blistein, Marcel and Ram, Ori and Zhang, Dan and Rosen, Evan and others},
  journal={arXiv preprint arXiv:2507.06261},
  year={2025}
}

@article{latif2025openaiO1,
  title   = {Comparative evaluation of OpenAI O1 and human performance in higher order cognition},
  author  = {Latif, Ehsan and Zhou, Yifan and Guo, Shuchen and Gao, Yizhu and Shi, Lehong and Nyaaba, Matthew and Bewerdorff, Arne and Yang, Xiantong and Zhai, Xiaoming and others},
  journal = {Scientific Reports},
  year    = {2025},
  doi     = {10.1038/s41598-025-33629-9}
}

@article{mei2025expo,
  title={ExPO-HM: Learning to Explain-then-Detect for Hateful Meme Detection},
  author={Mei, Jingbiao and Sun, Mingsheng and Chen, Jinghong and Qin, Pengda and Li, Yuhong and Chen, Da and Byrne, Bill},
  journal={arXiv preprint arXiv:2510.08630},
  year={2025}
}

@article{bai2025qwen3vltechnicalreport,
  title   = {Qwen3-VL Technical Report},
  author  = {Bai, Shuai and Cai, Yuxuan and Chen, Ruizhe and Chen, Keqin and Chen, Xionghui and Cheng, Zesen and Deng, Lianghao and Ding, Wei and Gao, Chang and Ge, Chunjiang and Ge, Wenbin and Guo, Zhifang and Huang, Qidong and Huang, Jie and Huang, Fei and Hui, Binyuan and Jiang, Shutong and Li, Zhaohai and Li, Mingsheng and Li, Mei and Li, Kaixin and Lin, Zicheng and Lin, Junyang and Liu, Xuejing and Liu, Jiawei and Liu, Chenglong and Liu, Yang and Liu, Dayiheng and Liu, Shixuan and Lu, Dunjie and Luo, Ruilin and Lv, Chenxu and Men, Rui and Meng, Lingchen and Ren, Xuancheng and Ren, Xingzhang and Song, Sibo and Sun, Yuchong and Tang, Jun and Tu, Jianhong and Wan, Jianqiang and Wang, Peng and Wang, Pengfei and Wang, Qiuyue and Wang, Yuxuan and Xie, Tianbao and Xu, Yiheng and Xu, Haiyang and Xu, Jin and Yang, Zhibo and Yang, Mingkun and Yang, Jianxin and Yang, An and Yu, Bowen and Zhang, Fei and Zhang, Hang and Zhang, Xi and Zheng, Bo and Zhong, Humen and Zhou, Jingren and Zhou, Fan and Zhou, Jing and Zhu, Yuanzhi and Zhu, Ke},
  journal = {arXiv preprint arXiv:2511.21631},
  year    = {2025},
  url     = {https://arxiv.org/abs/2511.21631}
}

@article{yang2025qwen3,
  title={Qwen3 technical report},
  author={Yang, An and Li, Anfeng and Yang, Baosong and Zhang, Beichen and Hui, Binyuan and Zheng, Bo and Yu, Bowen and Gao, Chang and Huang, Chengen and Lv, Chenxu and others},
  journal={arXiv preprint arXiv:2505.09388},
  year={2025}
}

@article{li2026qwen3,
  title={Qwen3-VL-Embedding and Qwen3-VL-Reranker: A Unified Framework for State-of-the-Art Multimodal Retrieval and Ranking},
  author={Li, Mingxin and Zhang, Yanzhao and Long, Dingkun and Chen, Keqin and Song, Sibo and Bai, Shuai and Yang, Zhibo and Xie, Pengjun and Yang, An and Liu, Dayiheng and others},
  journal={arXiv preprint arXiv:2601.04720},
  year={2026}
}

@article{pandiani2025toxic,
  title={‘Toxic’memes: A survey of computational perspectives on the detection and explanation of meme toxicities},
  author={Pandiani, Delfina S Martinez and Sang, Erik Tjong Kim and Ceolin, Davide},
  journal={Online Social Networks and Media},
  volume={47},
  pages={100317},
  year={2025},
  publisher={Elsevier}
}

@article{alafnan2025role,
  title={The role of memes in shaping political discourse on social media},
  author={AlAfnan, Mohammad Awad},
  journal={Studies in Media and Communication},
  volume={13},
  number={2},
  pages={1--10},
  year={2025},
  publisher={Redfame publishing}
}

@article{sharma2022detecting,
  title={Detecting and understanding harmful memes: A survey},
  author={Sharma, Shivam and Alam, Firoj and Akhtar, Md Shad and Dimitrov, Dimitar and Martino, Giovanni Da San and Firooz, Hamed and Halevy, Alon and Silvestri, Fabrizio and Nakov, Preslav and Chakraborty, Tanmoy},
  journal={arXiv preprint arXiv:2205.04274},
  year={2022}
}

@article{schmid2025humorous,
  title={Humorous hate speech on social media: A mixed-methods investigation of users’ perceptions and processing of hateful memes},
  author={Schmid, Ursula Kristin},
  journal={New Media \& Society},
  volume={27},
  number={3},
  pages={1588--1606},
  year={2025},
  publisher={Sage Publications Sage UK: London, England}
}

@article{mihuailescu2024never,
  title={Never Mess With the “Memers”: How Meme Creators Are Redefining Contemporary Politics},
  author={Mih{\u{a}}ilescu, Mihaela-Georgiana},
  journal={Social Media+ Society},
  volume={10},
  number={4},
  pages={20563051241296256},
  year={2024},
  publisher={Sage Publications Sage UK: London, England}
}

@inproceedings{amalia2018meme,
  title={Meme opinion categorization by using optical character recognition (OCR) and na{\"\i}ve bayes algorithm},
  author={Amalia, Amalia and Sharif, Amer and Haisar, Fikri and Gunawan, Dani and Nasution, Benny B},
  booktitle={2018 third international conference on informatics and computing (ICIC)},
  pages={1--5},
  year={2018},
  organization={IEEE}
}

@inproceedings{boinepelli2020sis,
  title={Sis@ iiith at semeval-2020 task 8: An overview of simple text classification methods for meme analysis},
  author={Boinepelli, Sravani and Shrivastava, Manish and Varma, Vasudeva},
  booktitle={Proceedings of the Fourteenth Workshop on Semantic Evaluation},
  pages={1190--1194},
  year={2020}
}

@inproceedings{shrestha2020nlp_uiowa,
  title={NLP\_UIOWA at SemEval-2020 Task 8: You’re not the only one cursed with knowledge-multi branch model memotion analysis},
  author={Shrestha, Ingroj and Rusert, Jonathan},
  booktitle={Proceedings of the Fourteenth Workshop on Semantic Evaluation},
  pages={891--900},
  year={2020}
}

@article{kiela2020hateful,
  title={The hateful memes challenge: Detecting hate speech in multimodal memes},
  author={Kiela, Douwe and Firooz, Hamed and Mohan, Aravind and Goswami, Vedanuj and Singh, Amanpreet and Ringshia, Pratik and Testuggine, Davide},
  journal={Advances in neural information processing systems},
  volume={33},
  pages={2611--2624},
  year={2020}
}

@inproceedings{lu2025having,
  title={Is Having Rationales Enough? Rethinking Knowledge Enhancement for Multimodal Hateful Meme Detection},
  author={Lu, Junyu and Xu, Bo and Zhang, Xiaokun and Zhu, Haohao and Wang, Kaichun and Yang, Liang and Lin, Hongfei},
  booktitle={Proceedings of the 48th International ACM SIGIR Conference on Research and Development in Information Retrieval},
  pages={559--569},
  year={2025}
}

@inproceedings{cao2024modularized,
  title={Modularized networks for few-shot hateful meme detection},
  author={Cao, Rui and Lee, Roy Ka-Wei and Jiang, Jing},
  booktitle={Proceedings of the ACM Web Conference 2024},
  pages={4575--4584},
  year={2024}
}

@misc{heebridging,
  title={Bridging Modalities: Enhancing Cross-Modality Hate Speech Detection with Few-Shot In-Context Learning (2024)},
  author={Hee, MS and Kumaresan, A and Lee, RKW}
}

@article{liu2025mind,
  title={MIND: A Multi-agent Framework for Zero-shot Harmful Meme Detection},
  author={Liu, Ziyan and Fan, Chunxiao and Lou, Haoran and Wu, Yuexin and Deng, Kaiwei},
  journal={arXiv preprint arXiv:2507.06908},
  year={2025}
}

@inproceedings{kmainasi2025memeintel,
    title = "{M}eme{I}ntel: Explainable Detection of Propagandistic and Hateful Memes",
    author = "Kmainasi, Mohamed Bayan  and
      Hasnat, Abul  and
      Hasan, Md Arid  and
      Shahroor, Ali Ezzat  and
      Alam, Firoj",
    editor = "Christodoulopoulos, Christos  and
      Chakraborty, Tanmoy  and
      Rose, Carolyn  and
      Peng, Violet",
    booktitle = "Proceedings of the 2025 Conference on Empirical Methods in Natural Language Processing",
    month = nov,
    year = "2025",
    address = "Suzhou, China",
    publisher = "Association for Computational Linguistics",
    url = "https://aclanthology.org/2025.emnlp-main.1539/",
    doi = "10.18653/v1/2025.emnlp-main.1539",
    pages = "30251--30267",
    ISBN = "979-8-89176-332-6"
}

@article{wu2024visionllm,
  title={Visionllm v2: An end-to-end generalist multimodal large language model for hundreds of vision-language tasks},
  author={Wu, Jiannan and Zhong, Muyan and Xing, Sen and Lai, Zeqiang and Liu, Zhaoyang and Chen, Zhe and Wang, Wenhai and Zhu, Xizhou and Lu, Lewei and Lu, Tong and others},
  journal={Advances in Neural Information Processing Systems},
  volume={37},
  pages={69925--69975},
  year={2024}
}

@inproceedings{sun2025multi,
  title={Multi-Modal Large Language Models are Effective Vision Learners},
  author={Sun, Li and Ahuja, Chaitanya and Chen, Peng and D'Zmura, Matt and Batmanghelich, Kayhan and Bontrager, Philip},
  booktitle={2025 IEEE/CVF Winter Conference on Applications of Computer Vision (WACV)},
  pages={8617--8626},
  year={2025},
  organization={IEEE}
}

@misc{wei2025sftsecondrlupt,
      title={First SFT, Second RL, Third UPT: Continual Improving Multi-Modal LLM Reasoning via Unsupervised Post-Training}, 
      author={Lai Wei and Yuting Li and Chen Wang and Yue Wang and Linghe Kong and Weiran Huang and Lichao Sun},
      year={2025},
      eprint={2505.22453},
      archivePrefix={arXiv},
      primaryClass={cs.CL},
      url={https://arxiv.org/abs/2505.22453}, 
}

@misc{kaelbling1996reinforcementlearningsurvey,
      title={Reinforcement Learning: A Survey}, 
      author={L. P. Kaelbling and M. L. Littman and A. W. Moore},
      year={1996},
      eprint={cs/9605103},
      archivePrefix={arXiv},
      primaryClass={cs.AI},
      url={https://arxiv.org/abs/cs/9605103}, 
}

@article{wu2025sailing,
  title={Sailing by the Stars: A Survey on Reward Models and Learning Strategies for Learning from Rewards},
  author={Wu, Xiaobao},
  journal={arXiv preprint arXiv:2505.02686},
  year={2025}
}

@article{ouyang2022training,
  title={Training language models to follow instructions with human feedback},
  author={Ouyang, Long and Wu, Jeffrey and Jiang, Xu and Almeida, Diogo and Wainwright, Carroll and Mishkin, Pamela and Zhang, Chong and Agarwal, Sandhini and Slama, Katarina and Ray, Alex and others},
  journal={Advances in neural information processing systems},
  volume={35},
  pages={27730--27744},
  year={2022}
}

@article{kaufmann2024survey,
  title={A survey of reinforcement learning from human feedback},
  author={Kaufmann, Timo and Weng, Paul and Bengs, Viktor and H{\"u}llermeier, Eyke},
  year={2024}
}

@article{schulman2017proximal,
  title={Proximal policy optimization algorithms},
  author={Schulman, John and Wolski, Filip and Dhariwal, Prafulla and Radford, Alec and Klimov, Oleg},
  journal={arXiv preprint arXiv:1707.06347},
  year={2017}
}

@article{rafailov2023direct,
  title={Direct preference optimization: Your language model is secretly a reward model},
  author={Rafailov, Rafael and Sharma, Archit and Mitchell, Eric and Manning, Christopher D and Ermon, Stefano and Finn, Chelsea},
  journal={Advances in neural information processing systems},
  volume={36},
  pages={53728--53741},
  year={2023}
}

@misc{deepseekai2025deepseekr1incentivizingreasoningcapability,
      title={DeepSeek-R1: Incentivizing Reasoning Capability in LLMs via Reinforcement Learning}, 
      author={DeepSeek-AI and Daya Guo and Dejian Yang and Haowei Zhang and Junxiao Song and Ruoyu Zhang and Runxin Xu and Qihao Zhu and Shirong Ma and Peiyi Wang and Xiao Bi and Xiaokang Zhang and Xingkai Yu and Yu Wu and Z. F. Wu and Zhibin Gou and Zhihong Shao and Zhuoshu Li and Ziyi Gao and Aixin Liu and Bing Xue and Bingxuan Wang and Bochao Wu and Bei Feng and Chengda Lu and Chenggang Zhao and Chengqi Deng and Chenyu Zhang and Chong Ruan and Damai Dai and Deli Chen and Dongjie Ji and Erhang Li and Fangyun Lin and Fucong Dai and Fuli Luo and Guangbo Hao and Guanting Chen and Guowei Li and H. Zhang and Han Bao and Hanwei Xu and Haocheng Wang and Honghui Ding and Huajian Xin and Huazuo Gao and Hui Qu and Hui Li and Jianzhong Guo and Jiashi Li and Jiawei Wang and Jingchang Chen and Jingyang Yuan and Junjie Qiu and Junlong Li and J. L. Cai and Jiaqi Ni and Jian Liang and Jin Chen and Kai Dong and Kai Hu and Kaige Gao and Kang Guan and Kexin Huang and Kuai Yu and Lean Wang and Lecong Zhang and Liang Zhao and Litong Wang and Liyue Zhang and Lei Xu and Leyi Xia and Mingchuan Zhang and Minghua Zhang and Minghui Tang and Meng Li and Miaojun Wang and Mingming Li and Ning Tian and Panpan Huang and Peng Zhang and Qiancheng Wang and Qinyu Chen and Qiushi Du and Ruiqi Ge and Ruisong Zhang and Ruizhe Pan and Runji Wang and R. J. Chen and R. L. Jin and Ruyi Chen and Shanghao Lu and Shangyan Zhou and Shanhuang Chen and Shengfeng Ye and Shiyu Wang and Shuiping Yu and Shunfeng Zhou and Shuting Pan and S. S. Li and Shuang Zhou and Shaoqing Wu and Shengfeng Ye and Tao Yun and Tian Pei and Tianyu Sun and T. Wang and Wangding Zeng and Wanjia Zhao and Wen Liu and Wenfeng Liang and Wenjun Gao and Wenqin Yu and Wentao Zhang and W. L. Xiao and Wei An and Xiaodong Liu and Xiaohan Wang and Xiaokang Chen and Xiaotao Nie and Xin Cheng and Xin Liu and Xin Xie and Xingchao Liu and Xinyu Yang and Xinyuan Li and Xuecheng Su and Xuheng Lin and X. Q. Li and Xiangyue Jin and Xiaojin Shen and Xiaosha Chen and Xiaowen Sun and Xiaoxiang Wang and Xinnan Song and Xinyi Zhou and Xianzu Wang and Xinxia Shan and Y. K. Li and Y. Q. Wang and Y. X. Wei and Yang Zhang and Yanhong Xu and Yao Li and Yao Zhao and Yaofeng Sun and Yaohui Wang and Yi Yu and Yichao Zhang and Yifan Shi and Yiliang Xiong and Ying He and Yishi Piao and Yisong Wang and Yixuan Tan and Yiyang Ma and Yiyuan Liu and Yongqiang Guo and Yuan Ou and Yuduan Wang and Yue Gong and Yuheng Zou and Yujia He and Yunfan Xiong and Yuxiang Luo and Yuxiang You and Yuxuan Liu and Yuyang Zhou and Y. X. Zhu and Yanhong Xu and Yanping Huang and Yaohui Li and Yi Zheng and Yuchen Zhu and Yunxian Ma and Ying Tang and Yukun Zha and Yuting Yan and Z. Z. Ren and Zehui Ren and Zhangli Sha and Zhe Fu and Zhean Xu and Zhenda Xie and Zhengyan Zhang and Zhewen Hao and Zhicheng Ma and Zhigang Yan and Zhiyu Wu and Zihui Gu and Zijia Zhu and Zijun Liu and Zilin Li and Ziwei Xie and Ziyang Song and Zizheng Pan and Zhen Huang and Zhipeng Xu and Zhongyu Zhang and Zhen Zhang},
      year={2025},
      eprint={2501.12948},
      archivePrefix={arXiv},
      primaryClass={cs.CL},
      url={https://arxiv.org/abs/2501.12948}, 
}

@misc{shao2024deepseekmathpushinglimitsmathematical,
      title={DeepSeekMath: Pushing the Limits of Mathematical Reasoning in Open Language Models}, 
      author={Zhihong Shao and Peiyi Wang and Qihao Zhu and Runxin Xu and Junxiao Song and Xiao Bi and Haowei Zhang and Mingchuan Zhang and Y. K. Li and Y. Wu and Daya Guo},
      year={2024},
      eprint={2402.03300},
      archivePrefix={arXiv},
      primaryClass={cs.CL},
      url={https://arxiv.org/abs/2402.03300}, 
}

@inproceedings{ranaldi2025multilingual,
  title={Multilingual Reasoning via Self-training},
  author={Ranaldi, Leonardo and Pucci, Giulia},
  booktitle={Proceedings of the 2025 Conference of the Nations of the Americas Chapter of the Association for Computational Linguistics: Human Language Technologies (Volume 1: Long Papers)},
  pages={11566--11582},
  year={2025}
}

@inproceedings{chen2025predicate,
  title={Predicate-Guided Generation for Mathematical Reasoning},
  author={Chen, Jiajun and Tam, Yik-Cheung},
  booktitle={Proceedings of the 2025 Conference on Empirical Methods in Natural Language Processing},
  pages={9097--9110},
  year={2025}
}

@inproceedings{cao2023pro,
  title={Pro-cap: Leveraging a frozen vision-language model for hateful meme detection},
  author={Cao, Rui and Hee, Ming Shan and Kuek, Adriel and Chong, Wen-Haw and Lee, Roy Ka-Wei and Jiang, Jing},
  booktitle={Proceedings of the 31st ACM international conference on multimedia},
  pages={5244--5252},
  year={2023}
}

@misc{kiela2021hatefulmemeschallengedetecting,
      title={The Hateful Memes Challenge: Detecting Hate Speech in Multimodal Memes}, 
      author={Douwe Kiela and Hamed Firooz and Aravind Mohan and Vedanuj Goswami and Amanpreet Singh and Pratik Ringshia and Davide Testuggine},
      year={2021},
      eprint={2005.04790},
      archivePrefix={arXiv},
      primaryClass={cs.AI},
      url={https://arxiv.org/abs/2005.04790}, 
}

@inproceedings{fersini-etal-2022-semeval,
    title = "{S}em{E}val-2022 Task 5: Multimedia Automatic Misogyny Identification",
    author = "Fersini, Elisabetta  and
      Gasparini, Francesca  and
      Rizzi, Giulia  and
      Saibene, Aurora  and
      Chulvi, Berta  and
      Rosso, Paolo  and
      Lees, Alyssa  and
      Sorensen, Jeffrey",
    editor = "Emerson, Guy  and
      Schluter, Natalie  and
      Stanovsky, Gabriel  and
      Kumar, Ritesh  and
      Palmer, Alexis  and
      Schneider, Nathan  and
      Singh, Siddharth  and
      Ratan, Shyam",
    booktitle = "Proceedings of the 16th International Workshop on Semantic Evaluation (SemEval-2022)",
    month = jul,
    year = "2022",
    address = "Seattle, United States",
    publisher = "Association for Computational Linguistics",
    url = "https://aclanthology.org/2022.semeval-1.74/",
    doi = "10.18653/v1/2022.semeval-1.74",
    pages = "533--549"
}

@article{pramanick2021momenta,
  title={MOMENTA: A multimodal framework for detecting harmful memes and their targets},
  author={Pramanick, Shraman and Sharma, Shivam and Dimitrov, Dimitar and Akhtar, Md Shad and Nakov, Preslav and Chakraborty, Tanmoy},
  journal={arXiv preprint arXiv:2109.05184},
  year={2021}
}

@inproceedings{alam-etal-2024-armeme,
    title = "{A}r{M}eme: Propagandistic Content in {A}rabic Memes",
    author = "Alam, Firoj  and
      Hasnat, Abul  and
      Ahmad, Fatema  and
      Hasan, Md. Arid  and
      Hasanain, Maram",
    editor = "Al-Onaizan, Yaser  and
      Bansal, Mohit  and
      Chen, Yun-Nung",
    booktitle = "Proceedings of the 2024 Conference on Empirical Methods in Natural Language Processing",
    month = nov,
    year = "2024",
    address = "Miami, Florida, USA",
    publisher = "Association for Computational Linguistics",
    url = "https://aclanthology.org/2024.emnlp-main.1173/",
    doi = "10.18653/v1/2024.emnlp-main.1173",
    pages = "21071--21090"
}

@inproceedings{mathias-etal-2021-findings,
    title = "Findings of the {WOAH} 5 Shared Task on Fine Grained Hateful Memes Detection",
    author = "Mathias, Lambert  and
      Nie, Shaoliang  and
      Mostafazadeh Davani, Aida  and
      Kiela, Douwe  and
      Prabhakaran, Vinodkumar  and
      Vidgen, Bertie  and
      Waseem, Zeerak",
    editor = "Mostafazadeh Davani, Aida  and
      Kiela, Douwe  and
      Lambert, Mathias  and
      Vidgen, Bertie  and
      Prabhakaran, Vinodkumar  and
      Waseem, Zeerak",
    booktitle = "Proceedings of the 5th Workshop on Online Abuse and Harms (WOAH 2021)",
    month = aug,
    year = "2021",
    address = "Online",
    publisher = "Association for Computational Linguistics",
    url = "https://aclanthology.org/2021.woah-1.21/",
    doi = "10.18653/v1/2021.woah-1.21",
    pages = "201--206"
}

@article{grasso2024kermit,
  title={Kermit: Knowledge-empowered model in harmful meme detection},
  author={Grasso, Biagio and La Gatta, Valerio and Moscato, Vincenzo and Sperl{\`\i}, Giancarlo},
  journal={Information Fusion},
  volume={106},
  pages={102269},
  year={2024},
  publisher={Elsevier}
}

@inproceedings{lin2024towards,
  title={Towards explainable harmful meme detection through multimodal debate between large language models},
  author={Lin, Hongzhan and Luo, Ziyang and Gao, Wei and Ma, Jing and Wang, Bo and Yang, Ruichao},
  booktitle={Proceedings of the ACM Web Conference 2024},
  pages={2359--2370},
  year={2024}
}

@inproceedings{xu2024exploring,
  title={Exploring chain-of-thought for multi-modal metaphor detection},
  author={Xu, Yanzhi and Hua, Yueying and Li, Shichen and Wang, Zhongqing},
  booktitle={Proceedings of the 62nd Annual Meeting of the Association for Computational Linguistics (Volume 1: Long Papers)},
  pages={91--101},
  year={2024}
}

@inproceedings{lin-etal-2023-beneath,
    title = "Beneath the Surface: Unveiling Harmful Memes with Multimodal Reasoning Distilled from Large Language Models",
    author = "Lin, Hongzhan  and
      Luo, Ziyang  and
      Ma, Jing  and
      Chen, Long",
    editor = "Bouamor, Houda  and
      Pino, Juan  and
      Bali, Kalika",
    booktitle = "Findings of the Association for Computational Linguistics: EMNLP 2023",
    month = dec,
    year = "2023",
    address = "Singapore",
    publisher = "Association for Computational Linguistics",
    url = "https://aclanthology.org/2023.findings-emnlp.611/",
    doi = "10.18653/v1/2023.findings-emnlp.611",
    pages = "9114--9128"
}

@article{kumari2024m3hop,
  title={M3hop-cot: Misogynous meme identification with multimodal multi-hop chain-of-thought},
  author={Kumari, Gitanjali and Jain, Kirtan and Ekbal, Asif},
  journal={arXiv preprint arXiv:2410.09220},
  year={2024}
}

@inproceedings{mathias2021findings,
  title={Findings of the WOAH 5 shared task on fine grained hateful memes detection},
  author={Mathias, Lambert and Nie, Shaoliang and Davani, Aida Mostafazadeh and Kiela, Douwe and Prabhakaran, Vinodkumar and Vidgen, Bertie and Talat, Zeerak},
  booktitle={Proceedings of the 5th Workshop on Online Abuse and Harms (WOAH 2021)},
  pages={201--206},
  year={2021}
}

@article{zhu2024distilling,
  title={Distilling mathematical reasoning capabilities into small language models},
  author={Zhu, Xunyu and Li, Jian and Liu, Yong and Ma, Can and Wang, Weiping},
  journal={Neural Networks},
  volume={179},
  pages={106594},
  year={2024},
  publisher={Elsevier}
}

@inproceedings{burbi2023mapping,
  title={Mapping memes to words for multimodal hateful meme classification},
  author={Burbi, Giovanni and Baldrati, Alberto and Agnolucci, Lorenzo and Bertini, Marco and Del Bimbo, Alberto},
  booktitle={Proceedings of the IEEE/CVF International Conference on Computer Vision},
  pages={2832--2836},
  year={2023}
}

@inproceedings{cao-etal-2022-prompting,
    title = "Prompting for Multimodal Hateful Meme Classification",
    author = "Cao, Rui  and
      Lee, Roy Ka-Wei  and
      Chong, Wen-Haw  and
      Jiang, Jing",
    editor = "Goldberg, Yoav  and
      Kozareva, Zornitsa  and
      Zhang, Yue",
    booktitle = "Proceedings of the 2022 Conference on Empirical Methods in Natural Language Processing",
    month = dec,
    year = "2022",
    address = "Abu Dhabi, United Arab Emirates",
    publisher = "Association for Computational Linguistics",
    url = "https://aclanthology.org/2022.emnlp-main.22",
    doi = "10.18653/v1/2022.emnlp-main.22",
    pages = "321--332",
}

@article{wu2024multimodal,
  title={Multimodal hateful meme classification based on transfer learning and a cross-mask mechanism},
  author={Wu, Fan and Chen, Guolian and Cao, Junkuo and Yan, Yuhan and Li, Zhongneng},
  journal={Electronics},
  volume={13},
  number={14},
  pages={2780},
  year={2024},
  publisher={MDPI}
}

@inproceedings{mei2024improving,
  title={Improving hateful meme detection through retrieval-guided contrastive learning},
  author={Mei, Jingbiao and Chen, Jinghong and Lin, Weizhe and Byrne, Bill and Tomalin, Marcus},
  booktitle={Proceedings of the 62nd Annual Meeting of the Association for Computational Linguistics (Volume 1: Long Papers)},
  pages={5333--5347},
  year={2024}
}

@inproceedings{yang2024uncertainty,
  title={Uncertainty-guided modal rebalance for hateful memes detection},
  author={Yang, Chuanpeng and Liu, Yaxin and Zhu, Fuqing and Han, Jizhong and Hu, Songlin},
  booktitle={Proceedings of the 62nd Annual Meeting of the Association for Computational Linguistics (Volume 1: Long Papers)},
  pages={4361--4371},
  year={2024}
}

@article{james1984estimating,
  title={Estimating within-group interrater reliability with and without response bias.},
  author={James, Lawrence R and Demaree, Robert G and Wolf, Gerrit},
  journal={Journal of applied psychology},
  volume={69},
  number={1},
  pages={85},
  year={1984},
  publisher={American Psychological Association}
}

@misc{zhang2020bertscoreevaluatingtextgeneration,
      title={BERTScore: Evaluating Text Generation with BERT}, 
      author={Tianyi Zhang and Varsha Kishore and Felix Wu and Kilian Q. Weinberger and Yoav Artzi},
      year={2020},
      eprint={1904.09675},
      archivePrefix={arXiv},
      primaryClass={cs.CL},
      url={https://arxiv.org/abs/1904.09675}, 
}

@inproceedings{banerjee-lavie-2005-meteor,
    title = "{METEOR}: An Automatic Metric for {MT} Evaluation with Improved Correlation with Human Judgments",
    author = "Banerjee, Satanjeev  and
      Lavie, Alon",
    editor = "Goldstein, Jade  and
      Lavie, Alon  and
      Lin, Chin-Yew  and
      Voss, Clare",
    booktitle = "Proceedings of the {ACL} Workshop on Intrinsic and Extrinsic Evaluation Measures for Machine Translation and/or Summarization",
    month = jun,
    year = "2005",
    address = "Ann Arbor, Michigan",
    publisher = "Association for Computational Linguistics",
    url = "https://aclanthology.org/W05-0909/",
    pages = "65--72"
}

\appendix

\appendix

\end{document}